\def\year{2021}\relax
\documentclass[letterpaper]{article} % DO NOT CHANGE THIS
\usepackage{aaai21}  % DO NOT CHANGE THIS
\usepackage{times}  % DO NOT CHANGE THIS
\usepackage{helvet} % DO NOT CHANGE THIS
\usepackage{courier}  % DO NOT CHANGE THIS
\usepackage[hyphens]{url}  % DO NOT CHANGE THIS
\usepackage{graphicx} % DO NOT CHANGE THIS
\usepackage{natbib}  % DO NOT CHANGE THIS AND DO NOT ADD ANY OPTIONS TO IT
\usepackage{caption} % DO NOT CHANGE THIS AND DO NOT ADD ANY OPTIONS TO IT
\usepackage{subfig}
\usepackage{paralist, tabularx}
\usepackage{newcommand}
\usepackage{adjustbox}
\usepackage{amssymb,amsmath,bm,color}
\usepackage{booktabs} % for professional tables
\usepackage{multirow}
\usepackage{multicol}

\newcommand{\xmark}{\times}
\usepackage{algorithm}
\usepackage{algorithmic}
\title{Self-Progressing Robust Training}
\author {
    Minhao Cheng,\textsuperscript{\rm 1,2}
    Pin-Yu Chen,\textsuperscript{\rm 2}
    Sijia Liu,\textsuperscript{\rm 2} 
    Shiyu Chang,\textsuperscript{\rm 2}
    Cho-Jui Hsieh,\textsuperscript{\rm 1}
    Payel Das\textsuperscript{\rm 2}\\
}
\affiliations {
    \textsuperscript{\rm 1} Department of Computer Science, UCLA \\
    \textsuperscript{\rm 2} IBM Research \\
    \{mhcheng,chohsieh\}@cs.ucla.edu, \{pin-yu.chen,sijia.liu,shiyu.chang,daspa\}@ibm.com
}
\begin{document}

\maketitle

\begin{abstract}
Enhancing model robustness under new and even adversarial environments is a crucial milestone toward building trustworthy machine learning systems. Current robust training methods such as adversarial training explicitly uses an ``attack'' (e.g., $\ell_{\infty}$-norm bounded perturbation) to generate adversarial examples during model training for improving adversarial robustness. In this paper, we take a different perspective and propose a new framework called SPROUT, self-progressing robust training. During model training, SPROUT progressively adjusts training label distribution via our proposed parametrized label smoothing technique, making training free of attack generation and more scalable. We also motivate SPROUT using a general formulation based on vicinity risk minimization, which includes many robust training methods as special cases.
Compared with state-of-the-art adversarial training methods (PGD-$\ell_\infty$ and TRADES) under $\ell_{\infty}$-norm bounded attacks and various invariance tests, SPROUT consistently attains superior performance and is more scalable to large neural networks. Our results shed new light on scalable, effective and attack-independent robust training methods.
\end{abstract}

\section{Introduction}

While deep neural networks (DNNs) have achieved unprecedented performance on a variety of datasets and across domains, developing better training algorithms that are capable of strengthening model robustness is the next crucial milestone toward trustworthy and reliable machine learning systems. In recent years, DNNs trained by standard algorithms (i.e., the natural models) are shown to be vulnerable to adversarial input perturbations \citep{biggio2013evasion,szegedy2013intriguing}. Adversarial examples crafted by designed input perturbations can easily cause erroneous decision making of natural models \citep{goodfellow2014explaining} and thus intensify the demand for robust training methods. 

State-of-the-art robust training algorithms are primarily based on the methodology of adversarial training \citep{goodfellow2014explaining,madry2017towards}, which calls specific attack algorithms to generate adversarial examples during model training for learning robust models. Albeit effective, these methods have the following limitations: (i) \textit{poor scalability} -- the process of generating adversarial examples incurs considerable computation overhead. For instance, our experiments show that, with the same computation resources, standard adversarial training (with 7 attack iterations per sample in every minibatch) of Wide ResNet on CIFAR-10 consumes 10 times more clock time per training epoch when compared with standard training; (ii) \textit{attack specificity} -- adversarially trained models are usually most effective against the same attack they trained on, and the robustness may not generalize well to other types of attacks \citep{tramer2019adversarial,kang2019testing}; (iii) \textit{preference toward wider network} --  adversarial training is more effective when the networks have sufficient capacity (e.g., having more neurons in network layers) \citep{madry2017towards}.

\begin{figure*}[t]
    \centering
    \begin{tabular}{cccc}
            %\hspace{-2mm}
        \subfloat[Natural]{\includegraphics[width=0.21\textwidth]{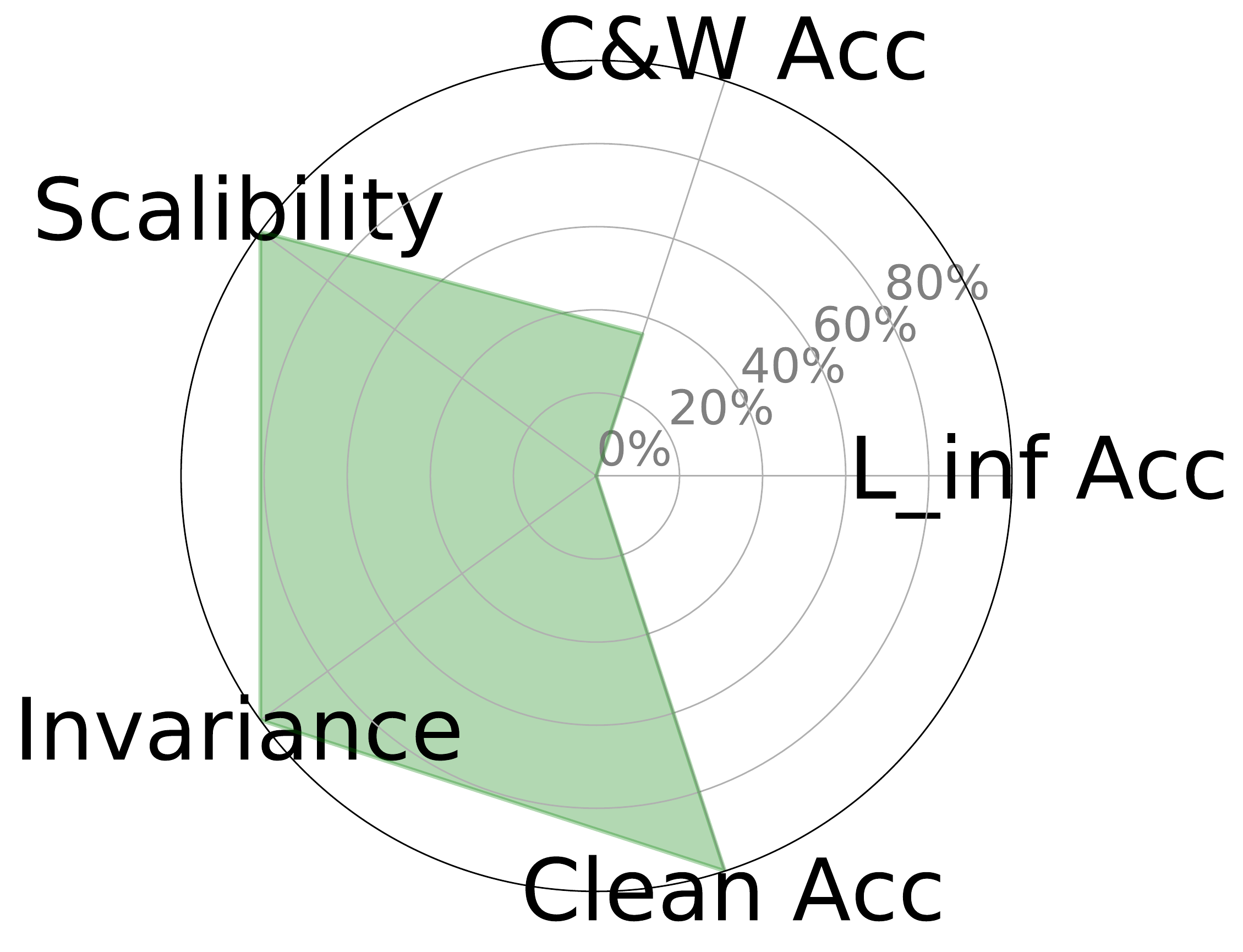}}
        %\hspace{-3mm}
		&
		\subfloat[Adversarial training]{\includegraphics[width=0.21\textwidth]{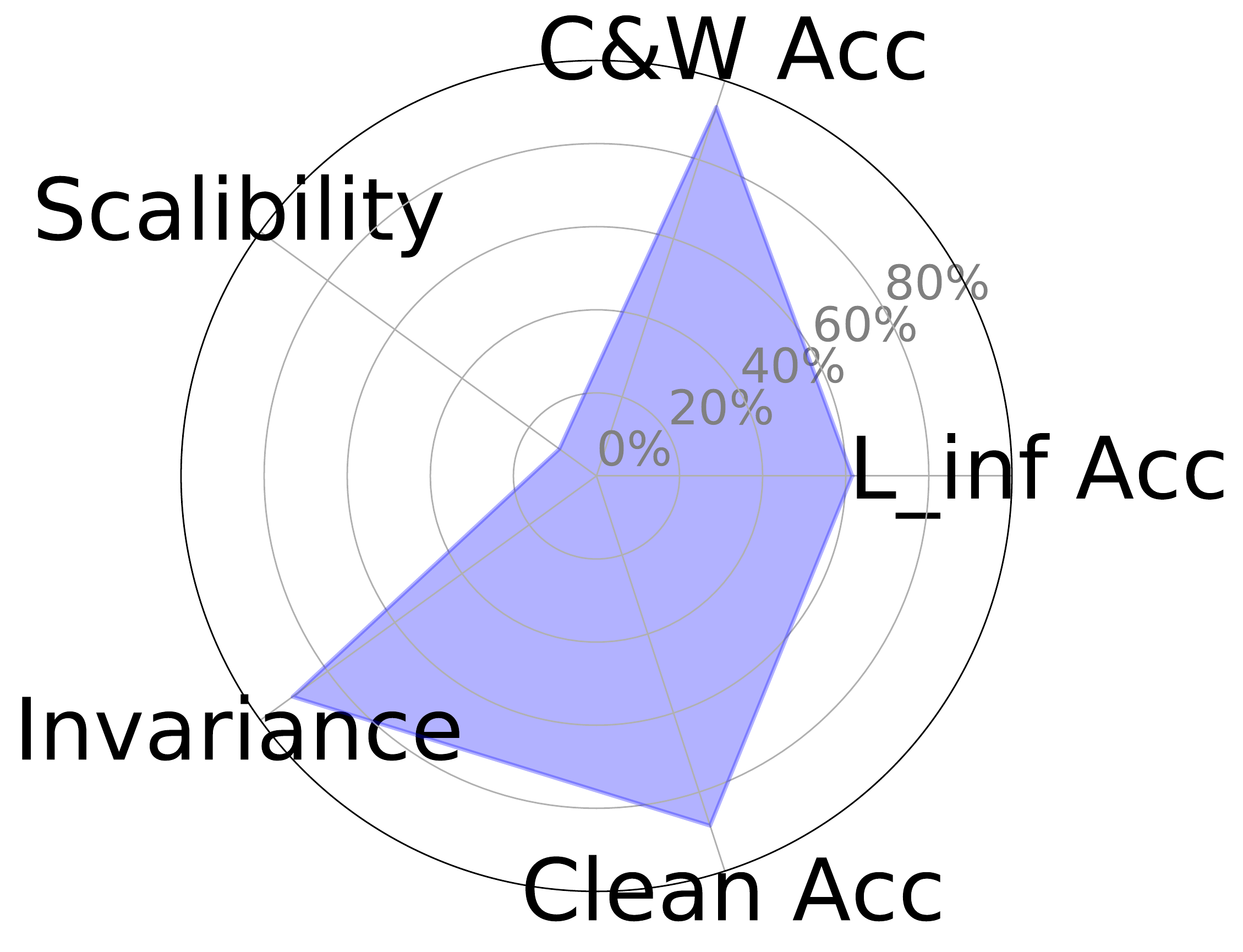}}
		%\hspace{-3mm}
		&
    \subfloat[TRADES]{\includegraphics[width=0.21\textwidth]{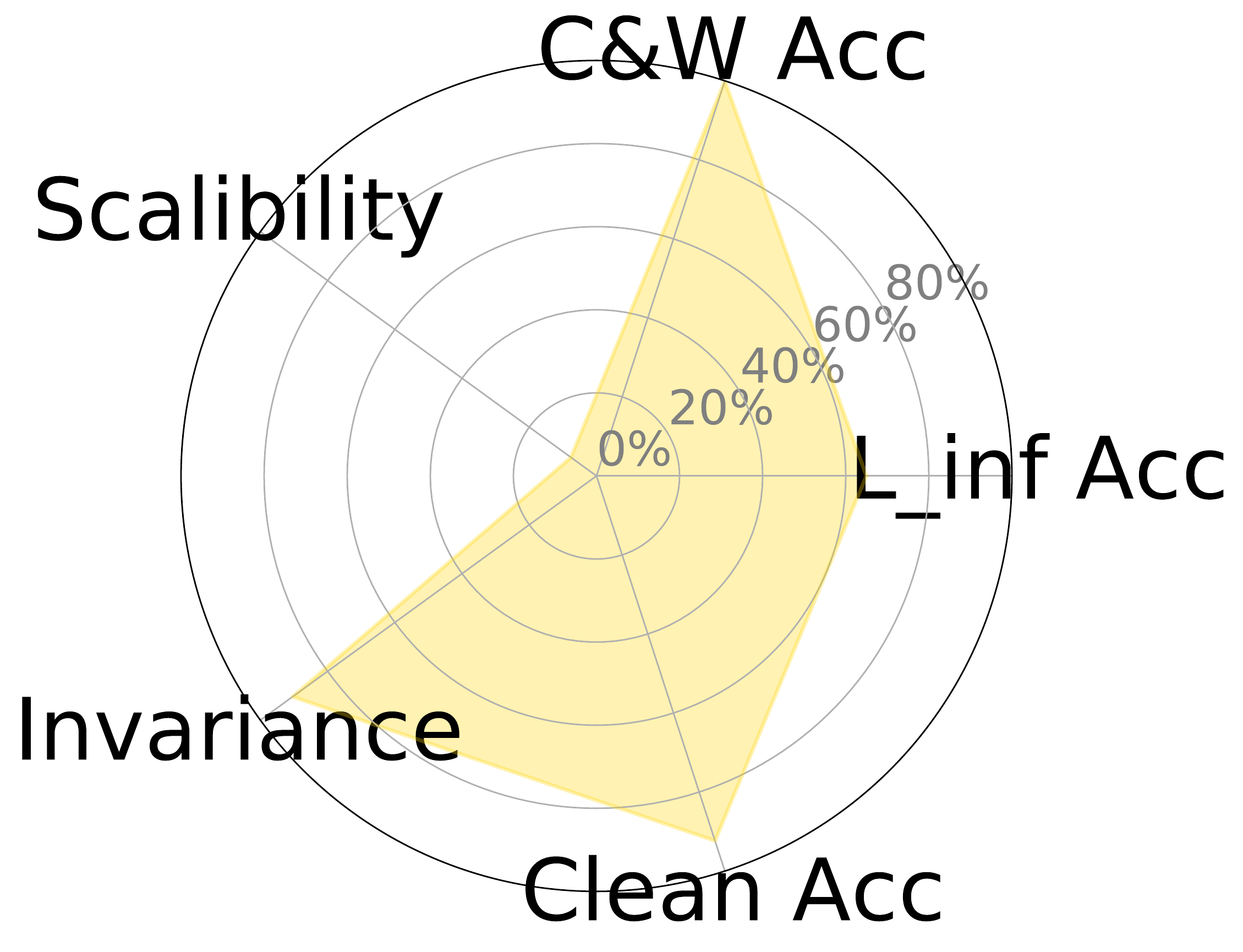}}
		 %\hspace{-3mm}
		&
		\subfloat[SPROUT (ours)]{\includegraphics[width=0.21\textwidth]{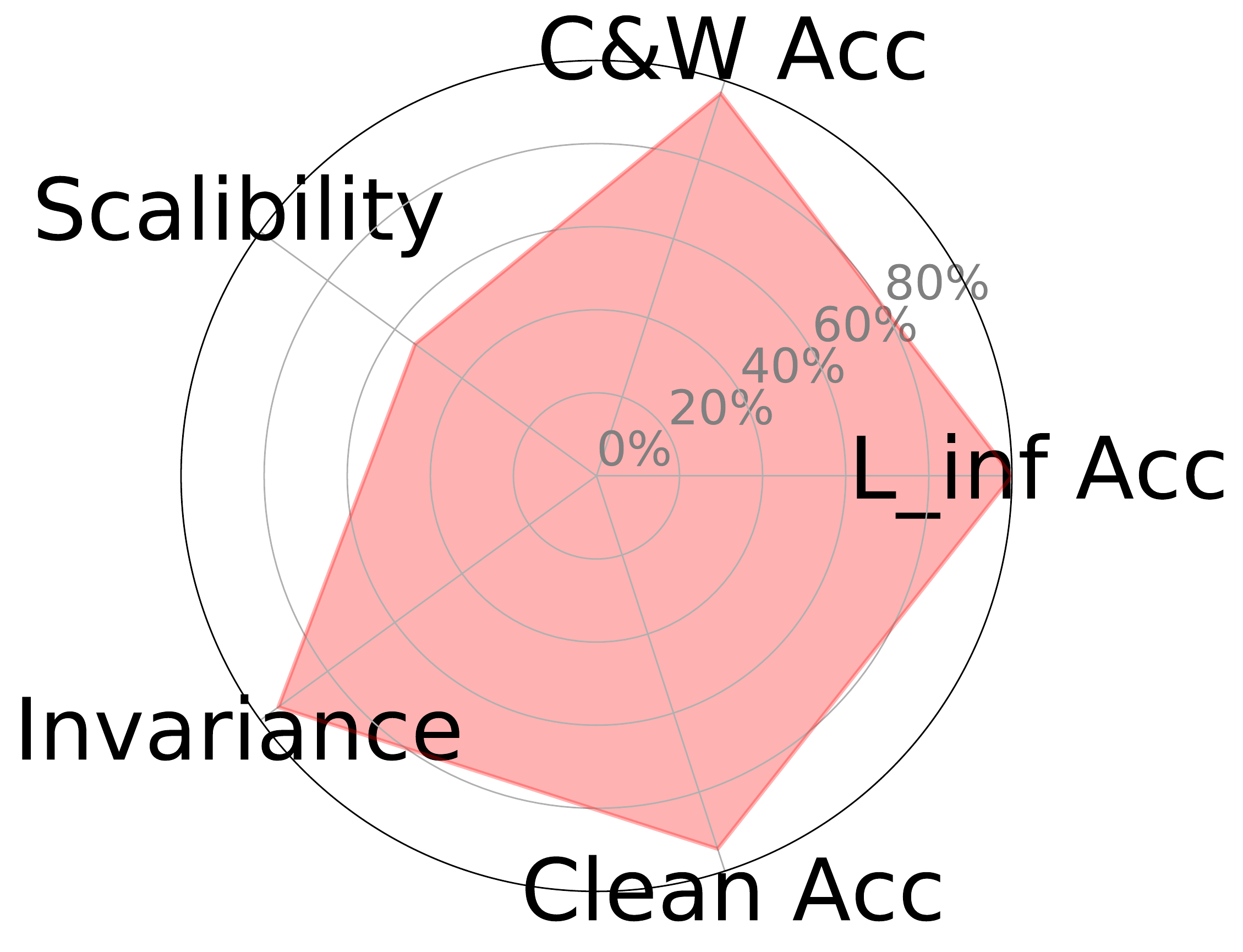}}

	\end{tabular}
		\caption{Multi-dimensional performance comparison of four training methods using VGG-16 network and CIFAR-10 dataset. 
		All dimensions are separately normalized by the best-performance method. The average score of each method is 0.6718 for natural (standard training), 0.6900 for PGD-$\ell_\infty$ based adversarial training \citep{madry2017towards}, 0.7107 for PGD-$\ell_\infty$ based TRADES \citep{zhang2019theoretically}, and 0.8798 for SPROUT (ours). The exact numbers are reported in Appendix. }
		\label{fig:radar}
		%\vspace{-2mm}
\end{figure*}

To address the aforementioned limitations, in this paper we propose a new robust training method named SPROUT, which is short for \underline{s}elf-\underline{p}rogressing \underline{r}\underline{o}b\underline{u}st \underline{t}raining. 
We motivate SPROUT by introducing a general framework that formulates robust training objectives via vicinity risk minimization (VRM), which includes many robust training methods as special cases. It is worth noting that the robust training methodology of SPROUT is fundamentally different from adversarial training, as SPROUT features self-adjusted label distribution during training instead of attack generation. In addition to our proposed parametrized label smoothing technique for progressive adjustment of training label distribution, SPROUT also adopts Gaussian augmentation and Mixup \citep{zhang2017mixup} to further enhance robustness. We show that they offer a complementary gain in robustness. 
In contrast to adversarial training, SPROUT spares the need for attack generation and thus makes its training scalable by a significant factor, while attaining better or comparable robustness performance on a variety of experiments. We also show exclusive features of SPROUT in terms of the novel findings that it can find robust models from either randomly initialized models or pretrained models, and its robustness performance is less sensitive to network width. Our implementation is publicly available \footnote{Code available at {\url{https://github.com/IBM/SPROUT}}}.

\subsection{Contributions}
\textbf{Multi-dimensional performance enhancement.} To illustrate the advantage of SPROUT over adversarial training and its variations, Figure \ref{fig:radar} compares the model performance of different training methods with the following five dimensions summarized from our experimental results: (i) \textit{Clean Acc} -- standard test accuracy, (ii) \textit{L\textunderscore inf Acc} -- accuracy under $\ell_{\infty}$-norm projected gradient descent (PGD) attack \citep{madry2017towards},
(iii)  \textit{C\&W Acc} -- accuracy under $\ell_2$-norm Carlini-Wagner (C\&W) attack, (iv) \textit{scalability} -- per epoch clock run-time, and (v) \textit{invariance} -- invariant transformation tests including rotation, brightness, contrast and gray images. Comparing to PGD-$\ell_\infty$ based adversarial training \citep{madry2017towards} and TRADES \citep{zhang2019theoretically}, SPROUT attains at least 20\% better L\textunderscore inf Acc, 2\% better Clean Acc, 5$\times$ faster run-time (scalability), 
2\% better invariance, while maintaining C\&W Acc, suggesting a new robust training paradigm that is scalable and comprehensive. 

%\PY{scalability number (change to secs). THis table will be moved to appendix}

% add a number table

We further summarize our main contributions as follows:
%\begin{enumerate}
\\
    $\bullet$ We propose SPROUT, a self-progressing robust training method composed of three modules that are efficient and free of attack generation: parametrized label smoothing, Gaussian augmentation, and Mixup. They altogether attain the state-of-the-art robustness performance and are scalable to large-scale networks.
    We will show that these modules are complementary to enhancing robustness.  We also perform an ablation study to demonstrate that our proposed parametrized label smoothing technique contributes to the major gain in boosting robustness. 
    \\
    $\bullet$ To provide technical explanations for SPROUT, we motivate its training methodology based on the framework of vicinity risk minimization (VRM). We show that many robust training methods, including attack-specific and attack-independent approaches, can be characterized as a specific form of VRM. The superior performance of SPROUT provides new insights on developing efficient robust training methods and theoretical analysis via VRM.
    \\
    $\bullet$ We evaluate the multi-dimensional performance of different training methods on (wide) ResNet and VGG networks using CIFAR-10 and ImageNet datasets. Notably, although SPROUT is attack-independent during training, we find that SPROUT significantly outperforms two major adversarial training methods, PGD-$\ell_\infty$ adversarial training \citep{madry2017towards} and TRADES \citep{zhang2019theoretically}, against the same type of attacks they used during training. Moreover, SPROUT is more scalable and runs at least 5$\times$ faster than adversarial training methods. It also attains higher clean accuracy, generalizes better to various invariance tests, and is less sensitive to network width.

\begin{table*}[t]
%\vspace{-2mm}
\caption{Summary of robust training methods using VRM formulation in \eqref{eqn:VRM_loss}. $\text{PGD}_\epsilon(\cdot)$ means (multi-step) PGD attack with perturbation budget $\epsilon$. $\text{Dirichlet}(\mathbf{b})$ is the Dirichlet distribution parameterized by $\mathbf{b}$. GA/LS stands for Gaussian-Augmentation/Label-Smoothing.
%\textcolor{blue}{ What is $p$ and $\tilde{p}$? Need to update Smooth training}
%\textcolor{green}{shall we explain column headers in the table caption?}
}
\label{tb:VRM}
%\vskip 0.15in
\begin{center}
\resizebox{!}{0.097\linewidth}{
\begin{tabular}{lcccccr}
\toprule
Methods & $g(\cdot)$ & $h(\cdot)$ & $\tilde{\bx}$ & $\tilde{\by}$ & attack-specific \\
\midrule
Natural & $\cI$ & $\cI$ & $\bx$& $\by$ & $\xmark$\\ 
GA \citep{zantedeschi2017efficient} & $\cI$ & $\cI$ & $\cN(\bx,\Delta^2)$ & $\by$ & $\xmark$\\
LS \citep{szegedy2016rethinking} & $\cI$ & $(1-\alpha)\by+\alpha \bu$ & $\bx$ & $\by$ &$\xmark$ \\
Adversarial training \citep{madry2017towards} & $\cI$ & $\cI$ & $\text{PGD}_\epsilon(\bx)$ & $\by$ & $\checkmark$\\
TRADES \citep{zhang2019theoretically} & $\cI$ & $(1-\alpha)\by+\alpha f(\tilde{\bx})$ & $\text{PGD}_\epsilon(\bx)$ & $\by$ & $\checkmark$\\
Stable training \citep{zheng2016improving} & $f(\bx)\circ f(\tilde{\bx})$ & $\cI$ & $\cN(\bx,\Delta^2)$ &$\by$ & $\xmark$\\
Mixup \citep{zhang2017mixup} & $\cI$ & $\cI$ & $(1-\lambda)\bx_i+\lambda\bx_j$ & $(1-\lambda)\by_i+\lambda\by_j$ &$\xmark$\\
LS+GA \citep{shafahi2019label} & $\cI$ & $(1-\alpha)\by+\alpha \bu$ & $\cN(\bx,\Delta^2)$ & $\by$ &$\xmark$\\
Bilateral Adv Training \citep{wang2018bilateral} & $\cI$ & $\cI$ & $\text{PGD}_\epsilon(\bx)$ (one or two step) &   $(1-\alpha)\by_i+\alpha \text{PGD}_{\epsilon^\prime}(\by)$ & $\checkmark$\\
SPROUT (ours) & $\cI$ & Dirichlet$((1-\alpha)\by+\alpha\bbeta)$ &  $(1-\lambda)\cN(\bx_i,\Delta^2)+\lambda \cN(\bx_j,\Delta^2)$ & $(1-\lambda)\by_i+\lambda \by_j$ &$\xmark$\\
\bottomrule
\end{tabular}
}
\end{center}
%\vspace{-2mm}
\end{table*}

\subsection{Related Work}
%\vspace{-4pt}
\textbf{Attack-specific robust training.} The seminal work of adversarial training with a first-order attack algorithm for generating adversarial examples \citep{madry2017towards} has greatly improved  adversarial robustness under the same threat model (e.g., $\ell_\infty$-norm bounded perturbations) as the attack algorithm. It has since inspired many advanced adversarial training algorithms with improved robustness. For instance, TRADES \citep{zhang2019theoretically} is designed to minimize a theoretically-driven upper bound on prediction error ofadversarial examples. \cite{liu2019rob} combined adversarial training with GAN to further improve robustness.
%which led to the first-ranked defense in the NeurIPS 2018 Adversarial Vision Challenge.
Bilateral adversarial training \citep{wang2018bilateral} finds robust models by adversarially perturbing the data samples and as well as the associated data labels. A feature-scattering based adversarial training method is proposed in \citep{zhang2019defense}.
Different from attack-specific robust training methods, our proposed SPROUT is free of attack generation, yet it can outperform attack-specific methods.
Another line of recent works uses an adversarially trained model along with additional unlabeled data \citep{carmon2019unlabeled,stanforth2019labels} or self-supervised learning with adversarial examples \citep{hendrycks2019using} to improve robustness, which in principle can also be used in SPROUT but is beyond the scope of this paper.

\textbf{Attack-independent robust training.} Here we discuss related works on Gaussian data augmentation, Mixup and label smoothing. 
Gaussian data augmentation during training is a commonly used baseline method to improve model robustness \citep{zantedeschi2017efficient}.
\cite{liu2018towards,liu2018adv,liu2019neural} demonstrated that additive Gaussian noise at both input and intermediate layers can  improve robustness.
 \cite{cohen2019certified} showed that Gaussian augmentation at the input layer can lead to certified robustness, which can also be incorporated in the training objective~\cite{zhai2020macer}.
 %scalable and certifiable defense method called random smoothing. 
Mixup \citep{zhang2017mixup} and its variants \citep{verma2018manifold,thulasidasan2019mixup}
are a recently proposed approach to improve model robustness and generalization by training a model on convex combinations of data sample pairs and their labels.
Label smoothing  was originally proposed in \citep{szegedy2016rethinking} as a regularizer to stabilize model training. The main idea is to replace one-hot encoded labels by assigning non-zero (e.g., uniform) weights to every label other than the original training label. Although label smoothing is also shown to benefit model robustness \citep{shafahi2019label,goibert2019adversarial}, its robustness gain is relatively marginal when compared to adversarial training. In contrast to currently used static (i.e., pre-defined) label smoothing functions, in SPROUT we propose a novel parametrized label smoothing scheme, which enables adaptive sampling of training labels from a parameterized distribution on the label simplex. The parameters of the label distribution are progressively adjusted according to the updates of model weights.

\section{General Framework for Formulating Robust Training}
%\vspace{-4pt}
\label{sec_VRM}
%\subsection{Vicinity Risk Minimization (VRM)}
The task of supervised learning is essentially learning a $K$-class classification function $f\in\cF$ that has a desirable mapping between a data sample $\bx \in \cX$ and the corresponding label $\by \in \cY$. Consider a loss function $L$ that penalizes the difference between the prediction $f(\bx)$ and the true label $\by$ from an unknown data distribution $P$, $(\bx,\by)\sim P$. The population risk can be expressed as 
\begin{equation}
    R(f)=\int L(f(\bx),\by) P(\bx,\by) d\bx d\by
\end{equation}
However, as the distribution $P$ is unknown, in practice machine learning uses empirical risk minimization (ERM) with the empirical data distribution of $n$ training data $\{x_i,y_i\}_{i=1}^n$
\begin{equation}
    P_\delta(\bx,\by)=\frac{1}{n}\sum_{i=1}^n\delta(\bx=\bx_i,\by=\by_i)
\end{equation}
to approximate $P(\bx,\by)$, where $\delta$ is a Dirac mass. Notably, a more principled approach is to use Vicinity Risk Minimization (VRM) \citep{chapelle2001vicinal}, defined as
\begin{equation}
    P_\nu(\bx,\by)=\frac{1}{n}\sum_{i=1}^n\nu(\tilde{\bx},\tilde{\by}|\bx_i,\by_i)
\end{equation}
where $\nu$ is a vicinity distribution that measures the probability of finding the virtual sample-label pair $(\tilde{\bx},\tilde{\by})$ in the vicinity of the training pair $(\bx_i,\by_i)$. Therefore, ERM can be viewed as a special case of VRM when $\nu=\delta$. VRM has also been used to motivate Mixup training \citep{zhang2017mixup}.  
Based on VRM, we propose a general framework that encompasses the objectives of many robust training methods as the following generalized cross entropy loss: 

\begin{equation}
\label{eqn:VRM_loss}
    H(\tilde{\bx},\tilde{\by},f) = -\sum_{k=1}^K [\log g(f(\tilde{\bx})_k)]h(\tilde{y}_k)
\end{equation}
where $f(\tilde{\bx})_k$ is the model's $k$-th class prediction probability on $\tilde{\bx}$, $g(\cdot): \RR \rightarrow \RR$ is a mapping adjusting the probability output, and $h(\cdot): \RR \rightarrow \RR$ is a mapping adjusting the training label distribution.
When $\tilde{\bx}=\bx$, $\tilde{\by}=\by$ and $g=h=\cI$, where $\cI$ denotes the identity mapping function, the loss in \eqref{eqn:VRM_loss} degenerates to the conventional cross entropy loss.

Based on the general VRM loss formulation in \eqref{eqn:VRM_loss}, in Table \ref{tb:VRM} we summarize a large body of robust training methods in terms of different expressions of $g(\cdot)$, $h(\cdot)$ and $(\tilde{\bx},\tilde{\by})$.  
%\citet{goodfellow2014explaining} firstly proposes adversarial training by feeding the adversarial examples generated by FGSM method back into the training and later 

For example, the vanilla adversarial training in \citep{madry2017towards} aims to minimize the loss of adversarial examples generated by the (multi-step) PGD attack with perturbation budget $\epsilon$, denoted by $\text{PGD}_\epsilon(\cdot)$. Its training objective can be rewritten as $\tilde{\bx}=\text{PGD}_\epsilon(\bx)$, $\tilde{\by}=\by$ and $g=h=\cI$. 
In addition to adversarial training only on perturbed samples of $\bx$, \citet{wang2018bilateral} designs adversarial label perturbation where it uses $\tilde{\bx}=\text{PGD}_\epsilon(\bx),~\tilde{\by}=(1-\alpha)\by+\alpha\text{PGD}_\epsilon(\by)$, and $\alpha \in [0,1]$ is a mixing parameter.
TRADES \citep{zhang2019theoretically} improves adversarial training with an additional regularization on the clean examples, 
%as well and show it has a closer bound the normal adversarial training, 
which is equivalent to replacing the label mapping function $h(\cdot)$ from identity to $(1-\alpha)\by+\alpha f(\tilde{\bx})$. %Since 
%the training procedure has to found the adversarial examples in each iteration, the scability of adversarial training is very poor. Also, \citet{madry2017towards} finds it requires a wider neural network to achieve good clean and robust accuracy at the same time. To overcome the limits of adversarial training,
Label smoothing (LS) alone is equivalent to the setup that $g=\cI$, $\tilde{\bx}=\bx$, $\tilde{\by}=\by$ and $h(\cdot)=(1-\alpha) \by + \alpha \bu$, where $\bu$ is often set as a uniform vector with value $1/K$ for a $K$-class supervised learning task.
Joint training with Gaussian augmentation (GA) and label smoothing (LS) as studied in
\citep{shafahi2019label} %finds the joint training with Gaussian augmentation and label smoothing could achieve a similar performance with adversarial training however training separately %does not help too much,
is equivalent to the case when $\tilde{\bx}=\cN(\bx,\Delta^2),\tilde{\by}=\by$, $g=\cI$ and $h(\by)=(1-\alpha)\by+\alpha/K$. We defer the connection between SPROUT and VRM to the next section.

%Therefore, these previous methods could be thought as a special choice of $g(\cdot)$ ,$h(\cdot)$, $\tilde{\bx}$ and $\tilde{\by}$ in our framework as listed in Table~\ref{tb:ref}. 

%\vspace{-4pt}
\section{SPROUT: Scalable Robust and Generalizable Training}
%\vspace{-4pt}
\label{section_SPROUT}
%\textcolor{blue}{we need to be more consistent when using boldface for vectors, especially for $y_i$}\\
In this section, we formally introduce SPROUT, a
novel robust training method that automatically finds a better vicinal risk function during training in a self-progressing manner.
%self-progressing robust training method that finding a better vicinal function automatically in the training process. 

%in order to both achieve a better clean accuracy without sacrificing clean accuracy, we propose SPROUT, a self-paced scalable training method, that finding a better vicinal function automatically in the training process.

%\paragraph{Motivation:} Adversarial training suffers from label leaking problem~\citep{}, which makes the trained model have a low clean accuracy. Although some remedies~\citep{} are proposed to address this problem, it is
%\vspace{-4pt}
\subsection{Self-Progressing Parametrized Label Smoothing}
%\vspace{-4pt}
To stabilize training and improve model generalization, \citet{szegedy2016rethinking} introduces label smoothing that converts “one-hot” label vectors into “one-warm” vectors representing low-confidence classification, in order to prevent a model from making over-confident predictions.
%Because large logit gaps produce high-confidence classifications, label-smoothed training data forces the classifier to produce small logit gaps. 
Specifically, the one-hot encoded label $\by$ is smoothed using
\begin{equation}
    \tilde{\by}=(1-\alpha)\by+\alpha \bu
\end{equation}
where $\alpha\in[0,1]$ is the smoothing parameter. A common choice is the uniform distribution $\bu=\frac{1}{K}$, where $K$ is the number of classes. 
%assumes each of Gaussian noise has the same adversarial robust effect and the label distribution other than the one-hot class is always the same, which ignores the fact the label itself has a strong correlation between each other. 
Later works \citep{wang2018bilateral,goibert2019adversarial} use an attack-driven label smoothing function $\bu$ to further improve adversarial robustness.
However, both uniform and attack-driven label smoothing disregard the inherent correlation between labels. 
%However, the correlation between classes is not considered. 
To address the label correlation, we propose to use the  Dirichlet distribution parametrized by $\bbeta \in \mathbb{R}^K_+$ 
for label smoothing.
Our SPROUT learns to update $\bbeta$ to find a training label distribution that is most uncertain to a given model $\theta$,
by solving
\begin{equation}
    \max_{\bbeta} L(\tilde{\bx},\tilde{\by},\bbeta;\theta)
\end{equation}
where $\tilde{\by} = \text{Dirichlet}((1-\alpha) \by+\alpha\bbeta)$. % means the labels are sampled from $\text{DirichLet}((1-\alpha) \by+\alpha\bbeta)$. 
Notably, instead of using a pre-defined or attack-driven function for $\bu$ in label smoothing, our Dirichlet approach automatically finds a label simplex by optimizing $\bbeta$. Dirichlet distribution indeed takes label correlation into consideration as its generated label $\bz=[z_1,\ldots,z_K]$ has the statistical properties
\begin{equation}
\label{eqn_corr}
    \EE[z_s]= \frac{\beta_s}{\beta_0},~ \text{Cov}[z_s,z_t]=\frac{-\beta_s\beta_t}{\beta_0^2(\beta_0+1)},~\sum_{s=1}^K z_s =1
\end{equation}
where $\beta_0=\sum_{k=1}^K\beta_k$ and $s,t \in \{1,\ldots,K\}$, $s \neq t$. 
%Therefore, it catches a good correlation between classes of label distribution by learning the optimal Dirichlet parameter $\bbeta$. 
Moreover,  one-hot label and uniform label smoothing are our special cases when $\bbeta=\by$ and $\bbeta=\bu$, respectively.  Our Dirichlet label smoothing co-trains with the update in model weights $\theta$ during training (see Algorithm \ref{alg:sprout}).
%In our implementation, Dirichlet label smoothing samples training labels from a Dirichlet distribution parametrized by $(1-\alpha)\by+\alpha \bbeta$, i.e., $\tilde{\by}=\text{Dirichlet}((1-\alpha)\by+\alpha \bbeta)$, where $\alpha \in [0,1]$ is the smoothing parameter.
% The advantage of our proposed self-progressing Dirichlet label smoothing over uniform label smoothing will be justified in our ablation study. In addition, we illustrate the label correlation learned from our Dirichlet label smoothing in Appendix.

%\textcolor{green}{I think illustrating the idea of updating Dirichlet distribution will be useful in terms of explaining as well as grabbing attention to the novelty of our work}
%\textcolor{blue}{mention iterative training}
%\vspace{-4pt}
\subsection{Gaussian Data Augmentation and Mixup}
%\vspace{-4pt}
\textbf{Gaussian augmentation.} Adding Gaussian noise to data samples during training is a common practice to improve model robustness. Its corresponding vicinal function is the Gaussian vicinity function $\nu(\tilde{\bx}_i,\tilde{\by}_i|\bx_i,\by_i)=\cN(\bx_i,\Delta^2)\delta(\tilde{\by}_i=\by_i)$, where  $\Delta^2$ is the variance of a standard normal random vector. However, the gain of Gaussian augmentation in robustness is marginal when compared with adversarial training (see our ablation study).  \citet{shafahi2019label} finds that combining uniform or attack-driven label smoothing with Gaussian smoothing can further improve adversarial robustness. Therefore, we propose to incorporate Gaussian augmentaion with Dirichlet label smoothing. 
 %However, it suffers from high dimensional problem. In order to secure $\epsilon$ robustness, the Gaussian noise added per example should be sufficient enough for the $\epsilon$ ball. Besides, Gaussian vicinities assumes the label distribution keeps the same, which causes the label leaking problem same with adversarial training. 
 %Therefore, finding a better label distribution on the Gaussian vicinities is the key to further improve the adversarial robustness. We found  When combing with the self-paced label smoothing, the adversarial robustness could be greatly improved. \citet{shafahi2019label} observe the similar effect as well.
 The joint vicinity function becomes $ \nu(\tilde{\bx}_i,\tilde{\by}_i|\bx_i,\by_i,\bbeta)=\cN(\bx_i,\Delta^2)\delta(\tilde{\by}_i=
\text{Dirichlet}((1-\alpha)\by_i+\alpha\bbeta)) $.
%  \begin{equation}
%  \label{eqn_GA_LS}
%      \nu(\tilde{\bx}_i,\tilde{\by}_i|\bx_i,\by_i,\bbeta)=\cN(\bx_i,\Delta^2)\delta(\tilde{\by}_i=
%      \text{Dirichlet}((1-\alpha)\by_i+\alpha\bbeta)) 
%  \end{equation}
Training with this  vicinity function means drawing labels from the Dirichlet distribution for the original data sample $\bx_i$ and its neighborhood characterized by Gaussian augmentation.

%Therefore, the model could learn the added Gaussian noise with its corresponding label distribution draw from parametrized Dirichlet distribution

%\subsection{Integrating Mixup}
%However, although we could find a better label distribution through parameterized label smoothing given a random Gaussian noise, the label leaking problem still exists, where there is no other information about correlation between the added noise and true label. Although \citep{} suggests replacing the true label with the most likely label predicted by the model, it is counter-intuitive in the training process. 

\textbf{Mixup.} To further improve model generalization, SPROUT also integrates Mixup \citep{zhang2017mixup} that performs convex combination on pairs of training data samples (in a minibatch) and their labels during training. The vicinity function of Mixup is 
%\citep{zhang2017mixup} proposes mixup that is a replacement the ERM with a linear vicinal distribution
%\begin{equation}
    $\nu(\tilde{\bx},\tilde{\by}|\bx_i,\by_i)=\delta(\tilde{\bx}=(1-\lambda) \bx_i+\lambda\bx_j, \tilde{\by}=(1-\lambda) \by_i+\lambda\by_j)$,
%\end{equation}
where $\lambda\sim \text{Beta}(a,a)$ is the mixing parameter drawn from the Beta distribution and $a>0$ is the shape parameter. The Mixup vicinity function can be generalized to multiple data sample pairs. 
Unlike Gaussian augmentation which is irrespective of the label (i.e., only adding noise to $\bx_i$), Mixup aims to augment data samples on the line segments of training data pairs and assign them convexly combined labels during training.

\textbf{Vicinity function of SPROUT.}
With the aforementioned techniques integrated in SPROUT, 
%including Dirichlet label smoothing, Gaussian augmentation and Mixup,
the overall vicinity function of SPROUT can be summarized as $\nu(\tilde{\bx},\tilde{\by}|\bx_i,\by_i,\bbeta) =\delta(\tilde{\bx} =\lambda \cN(\bx_i,\Delta^2)+(1-\lambda)\cN(\bx_j,\Delta^2),~\tilde{\by}= \text{Dirichlet}((1-\alpha)((1-\lambda) \by_i+\lambda\by_j)+\alpha\bbeta)$.
%we propose our final SPROUT using the vicinal function as follows:
%\textcolor{blue}{The following equation needs to be revised}
% \begin{align}
% \label{VRM_sprout}
%     &\nu(\tilde{\bx},\tilde{\by}|\bx_i,\by_i,\bbeta) \\ &=\delta(\tilde{\bx} =\lambda \cN(\bx_i,\Delta^2)+(1-\lambda)\cN(\bx_j,\Delta^2), \tilde{\by}= \text{Dirichlet}((1-\alpha)((1-\lambda) \by_i+\lambda\by_j)+\alpha\bbeta) \nonumber
% \end{align}
%And the detailed learning procedure is showed in Algorithm \ref{alg:sprout}.

In the experiment, we will show that Dirichlet label smoothing, Gaussian augmentation and Mixup are complementary to enhancing robustness by showing their diversity in input gradients.

\begin{algorithm}[t]
  \caption{SPROUT algorithm}
  \label{alg:sprout}
\begin{algorithmic}
  \STATE {\bfseries Input:} Training dataset $(X,Y)$, Mixup parameter $\lambda$,
  Gaussian augmentation variance $\Delta^2$, model learning rate $\gamma_\theta$, Dirichlet label smoothing learning rate $\gamma_{\bbeta}$ and parameter $\alpha$, generalized cross entropy loss $L$
  \STATE Initial model $\theta$: random initialization (train from scratch) or pre-trained model checkpoint
  %\STATE \textcolor{blue}{what about the initialization of $\bbeta$?}
  \STATE Initial $\bbeta$: random initialization
  \FOR{epoch=$1,\dots,N$}
  \FOR{minibatch $X_B\subset X, Y_B \subset Y$}
  \STATE $X_B \leftarrow \cN(X_B,\Delta^2)$
  \STATE $X_{mix}, Y_{mix}$ $\leftarrow$ Mixup($X_B,Y_B,\lambda$)
  \STATE $Y_{mix} \leftarrow \text{Dirichlet}(\alpha Y_{mix}+(1-\alpha)\bbeta)$
  %{\color{red}{Need to update this part}}
  \STATE $g_\theta \leftarrow \nabla_\theta L(X_{mix},Y_{mix},\theta)$
  \STATE $g_{\bbeta} \leftarrow \nabla_{\bbeta} L(X_{mix},Y_{mix},\theta)$
  \STATE $\theta \leftarrow \theta - \gamma_\theta g_\theta$ %\textcolor{blue}{check $L$, can we use adaptive optimization here?}
  \STATE $\bbeta \leftarrow \bbeta + \gamma_{\bbeta} g_{\bbeta}$
  \ENDFOR
  \ENDFOR
  \STATE {\bfseries return} $\theta$
\end{algorithmic}
\end{algorithm}

\subsection{SPROUT Algorithm}

\label{subsec_algorithm}
Using the VRM framework, the training objective of SPROUT is
\begin{align}
    \min_{\theta}  \max_{\bbeta} \sum_{i=1}^n L (\nu(\tilde{\bx_i},\tilde{\by_i}|\bx_i,\by_i,\bbeta);\theta),
\end{align}
where $\theta$ denotes the model weights, $n$ is the number of training data, $L$ is the generalized cross entropy loss defined in \eqref{eqn:VRM_loss} and $\nu(\tilde{\bx},\tilde{\by}|\bx_i,\by_i,\bbeta)$ is the vicinity function of SPROUT.
Our SPROUT algorithm co-trains $\theta$ and $\bbeta$ via  stochastic gradient descent/ascent to solve the outer minimization problem on $\theta$ and the inner maximization problem on $\bbeta$.
In particular, for calculating the gradient $g_{\bbeta}$ of the  parameter $\bbeta$, we use the Pytorch implementation based on \citep{figurnov2018implicit}.
SPROUT can either train a model from scratch with randomly initialized $\theta$ or strengthen a pre-trained model. 
%when evaluated against PGD-$\ell_\infty$ attack with different $\epsilon$ perturbation constraints, 
We find that training from either randomly initialized or pre-trained natural models using SPROUT can yield substantially robust models that are resilient to large perturbations (see Appendix).
The training steps of SPROUT are summarized in Algorithm \ref{alg:sprout}. 
%\textcolor{blue}{How do we obtain the gradient wrt $\bbeta$? Add a ref or illustration }

We also note that our min-max training methodology is different from the min-max formulation in adversarial training \citep{madry2017towards}, which is  $\min_{\theta} \sum_{i=1}^n  \max_{\boldsymbol{\delta}_i: \|\boldsymbol{\delta}_i\|_p \leq \epsilon} L(\bx_i+\boldsymbol{\delta}_i,\by_i;\theta)$, where $\|\boldsymbol{\delta}_i\|_p$ denotes the $\ell_p$ norm of the adversarial perturbation $\boldsymbol{\delta}_i$. While the outer minimization step for optimizing $\theta$ can be identical, the inner maximization of adversarial training requires running multi-step PGD attack to find adversarial perturbations $\{\boldsymbol{\delta}_i\}$ for each data sample in every minibatch (iteration) and epoch, which is attack-specific and time-consuming (see our scalability analysis in Table \ref{tab:scalability}). On the other hand, our inner maximization is upon the Dirichlet parameter $\bbeta$, which is attack-independent and only requires single-step stochastic gradient ascent with a minibatch to update $\bbeta$. We have explored multi-step stochastic gradient ascent on $\bbeta$ and found no significant performance enhancement but increased computation time.

\textbf{Advantages of SPROUT.} Comparing to adversarial training, the training of SPROUT is more efficient and scalable, as it only requires one additional back propagation to update $\bbeta$ in each iteration (see  Table \ref{tab:scalability} for a run-time analysis). As highlighted in Figure \ref{fig:radar}, SPROUT is also more comprehensive as it automatically improves robustness in multiple dimensions owing to its self-progressing training methodology. Moreover, we find that SPROUT significantly outperforms adversarial training and attains larger gain in robustness as
network width increases (see Figure \ref{fig:width}), which makes SPROUT a promising approach to support robust training for a much larger set of network architectures.

\section{Performance Evaluation}
%\vspace{-4pt}
\subsection{Experiment Setup}
\textbf{Dataset and network structure.} We use CIFAR-10~\citep{cifar10} and ImageNet~\citep{deng2009imagenet} for performance evaluation. For CIFAR-10, we use both standard VGG-16~\citep{simonyan2014very} and Wide ResNet. The Wide ResNet models are pre-trained PGD-$\ell_\infty$ robust models given by adversarial training and TRADES. For VGG-16, we
implement adversarial training with the standard hyper-parameters and train TRADES using the official implementation.
For ImageNet, we use ResNet-152.   
%For ImageNet, we use ResNet-50
%For Wide ResNet, we use the models released by adversarial training and TRADES official repository and implement the Madry's adversarial training using the standard hyper-parameter. 
All our experiments were implemented in Pytorch-1.2 
%and publicly available at {\color{blue\url{}}}. 
and conducted using dual Intel E5-2640 v4 CPUs (2.40GHz) with 512 GB memory with a GTX 1080 GPU. %\textcolor{blue}{mention CPU, RAM, GPU as well}

\textbf{Implementation details.} 
As suggested in Mixup \citep{zhang2017mixup},  we set the Beta distribution parameter $a=0.2$ when sampling the mixing parameter $\lambda$. For Gaussian augmentation, we set $\Delta=0.1$, which is within the suggested range in \citep{zantedeschi2017efficient}. Also, we set the label smoothing parameter $\alpha=0.01$. %\PY{CHECK $\alpha$ again}
A parameter sensitivity analysis on $\lambda$ and $\alpha$ is given in Appendix. 
%\PY{how about $\alpha$ in Dirichlet LS?}
Unless specified otherwise, for SPROUT we set the model initialization to be a natural model. An ablation study of model initialization is given in ablation study. 
% Our implementation is publicly available at \textcolor{blue}{\url{https://github.com/IBM/SPROUT}}.
%Also, we found SPROUT benefits on training from a natural trained model, while the other methods don't. 
%Therefore, we training from a natural pre-trained model using SPROUT and training all the other methods from scratch or using the provided checkpoint. 
%\vspace{-4pt}

% \begin{figure*}[htbp]
%     \centering
%     \begin{tabular}{cccc}
%         \subfloat{\includegraphics[width=0.23\textwidth]{imgs/vgg_20.pdf}}
% 		&
% 		\subfloat{\includegraphics[width=0.23\textwidth]{imgs/vgg_100.pdf}}
%         %\\ 
%         \hspace{ 2mm}
%         \subfloat{\includegraphics[width=0.23\textwidth]{imgs/wide_20.pdf}}
% 		&
% 		\subfloat{\includegraphics[width=0.23\textwidth]{imgs/wide_100.pdf}}
% 	\end{tabular}
% 	\vspace{-4mm}
% 		\caption{Robust accuracy of CIFAR-10 under PGD-$\ell_\infty$ attack. SPROUT significantly outperforms other methods.}
% 		\label{fig:res}
% 			\vspace{-4mm}
% \end{figure*}

% \begin{figure}[htbp]
%     \centering
%     \begin{tabular}{cc}
%         \subfloat{\includegraphics[width=0.23\textwidth]{imgs/vgg_cw.pdf}}
% 		&
% 		\subfloat{\includegraphics[width=0.23\textwidth]{imgs/wide_cw.pdf}}
% 	\end{tabular}
% 		\caption{Robust accuracy of CIFAR-10 under C\&W-$\ell_2$ attack} %\PY{need to fix the x-axis to \& }}
% 		\label{fig:cw}
% \end{figure}

\subsection{Adversarial Robustness under Various Attacks}

\label{subsec_robustness_attack}
\textbf{White-box attacks.}
On CIFAR-10, we compare the model accuracy under $\epsilon = 0.03 \approx 8/255$ strength of white-box $\ell_\infty$-norm bounded non-targeted PGD attack, which is considered as the strongest first-order adversary \citep{madry2017towards} with an $\ell_\infty$-norm constraint $\epsilon$  (normalized between 0 to 1).
%could consider as one of the strongest adversarial attack in $\ell_\infty$ constraint. 
All PGD attacks are implemented with random starts and we run PGD attack with 20, 100 and 200 steps in our experiments.
To be noted, we use both $\text{PGD}^{X}$ ($X$-step PGD with step size $\epsilon/5)$. As suggested, we test our model under different steps PGD and multiple random restarts. In Table \ref{tb:vgg}, we find SPROUT achieves 62.24\% and 66.23\% robust accuracy on VGG16 and Wide ResNet respectively, %{\color{red}(Cho: under what $\epsilon$?)}, 
while TRADES and adversarial training are 10-20\% worse than SPROUT. 
%We set the maximum (normalized) $\ell_\infty$ distortion from $[0:0.07:0.005]$ and reports their corresponding testing accuracy. 
% 
% The results are consistent with the vallina PGD-$\ell_\infty$ attack, i.e., SPROUT attains the highest accuracy (see Appendix \ref{appen_PGD_multi}). 
Moreover, we report the results of C\&W-$\ell_\infty$ attack \citep{carlini2017towards} in Appendix.
Next, we compare against $\ell_2$-norm based C\&W attack by using the default attack setting with 10 binary search steps and 1000 iterations per step to find successful perturbations while minimizing their $\ell_2$-norm. SPROUT can achieve 85.21\% robust accuracy under $\ell_2$ $\epsilon=0.05$ constraint while Adv train and TRADES achieves 77.76\% and 82.58\% respectively. It verifies that SPROUT can improve $\ell_\infty$ robustness by a large margin without degrading $\ell_2$ robustness.
SPROUT's accuracy under C\&W-$\ell_2$ attack is similar to TRADES and is better than both natural and adversarial training. The results also suggest that the attack-independent and self-progressing training nature of SPROUT can prevent the drawback of failing to provide comprehensive robustness to multiple and simultaneous $\ell_p$-norm attacks in adversarial training \citep{tramer2019adversarial,kang2019testing}.

\begin{table*}[tb]
\caption{The clean and robust accuracy of VGG-16  and Wide-ResNet 20 models trained by various defense methods. All robust accuracy results use $\epsilon=0.03$ ( $\ell_\infty$  perturbation). $B$ PGD$^{A}$ denotes an $A$-step PGD attack with $B$ random restarts.}
\label{tb:vgg}
\begin{center}
\resizebox{1.0\textwidth}{!}{
\begin{tabular}{l|c|c|c|c|c|c|c|c|c|c}
\toprule

&\multicolumn{5}{c|}{VGG-16} &  \multicolumn{5}{c}{Wide-ResNet 20}\\
\cline{2-11}
Methods &  No attack  & PGD$^{20}$ & PGD$^{100}$ & PGD$^{200}$& 10 PGD$^{100}$ & No attack  & PGD$^{20}$ & PGD$^{100}$ & PGD$^{200}$& 10 PGD$^{100}$\\
\hline
Natural train & {\bf 93.34\%} &  0.6\% & 0.1\% & 0.0\% & 0.0\% & {\bf 95.93\%} &  0.0\% & 0.0\% & 0.0\% & 0.0\%\\
Adv train \citep{madry2017towards} & 80.32\% & 36.63\% & 36.29\% &36.01\% &36.8\% & 87.25\% & 45.91\% & 45.32\% &45.02\% &44.98\% \\
TRADES \citep{zhang2019theoretically} &84.85\% & 38.81\% & 38.21\% &37.95\%  & 37.94\% &84.92\% & 56.23\% & 56.13\% &55.96\%  & 56.01\%  \\
SPROUT (ours) & 89.15\% & {\bf 62.24\%} & {\bf 58.93\%} & {\bf 57.9\%} &{\bf 58.08\%} & 90.56\% & {\bf 66.23\%} & {\bf 64.58\%} & {\bf 64.30\%} &{\bf 64.32\%}\\
\bottomrule
\end{tabular}
}
\end{center}
\end{table*}

% \subsection{Performance with different number of random starts for PGD attack}
% As suggested by \citep{madry2017towards}, PGD attack with multiple random starts is a stronger attack method to evaluate robustness. Therefore, in Table~\ref{tab:start}, we conduct the following experiment on CIFAR-10 and wide ResNet to show that the model trained by SPROUT can still attain at least 61\% accuracy against PGD-$\ell_{\infty}$ attack ($\epsilon=0.03$) with the number of random starts varying from 1 to 10 and with 20 attack iterations. The robust accuracy of SPROUT is still clearly higher than other methods as shown in Figure \ref{fig:res}. We also perform two additional attack settings: (i)
% 100-step PGD-$\ell_{\infty}$ attack with 10 random restarts using the CW loss and $\epsilon=0.03$; (ii) 100-step PGD-$\ell_{\infty}$ attack with 10 random restarts using the cross entropy and $\epsilon=0.03$. We find that SPROUT can still achieve 51.23\% robust accuracy in setting (i) and  61.18\% robust accuracy in setting (ii).

% \begin{table}[h]
%     \caption{Robust accuracy of SPROUT on  PGD-$\ell_\infty$ attack with $\epsilon=0.03$ using different number of random starts.}
%     \label{tab:start}
%     \centering
%     \begin{tabular}{c|c|c|c|c|c}
%     \toprule
%         $\#$ random start &1 & 3 & 5 & 8 & 10 \\
%          \hline
%         Robust accuracy & 64.58\% & 62.53\% & 61.98\% & 61.38\%& 61.00\%\\
%     \bottomrule
%     \end{tabular}
% \end{table}

\textbf{Transfer attack.}
We follow the criterion of evaluating transfer attacks in \citep{athalye2018obfuscated} to inspect whether the models trained by SPROUT will cause the issue of obfuscated gradients and give a false sense of robustness.
%rule out the model trained by SPROUT cause obfuscated gradient, we conduct black-box transfer attack experiments.
%In the experiments, 
We generate 10,000 PGD-$\ell_\infty$ adversarial examples from CIFAR-10 natural models with $\epsilon=0.03$ and evaluate their attack performance on the target model. 
Table \ref{tab:transfer} shows SPROUT achieves the best accuracy when compared with adversarial training and TRADES, suggesting the effectiveness of SPROUT in defending both white-box and transfer attacks.

\begin{table}[t]
%\vspace{-2mm}
    \caption{Robust accuracy of CIFAR-10 under transfer attack
    %The adversarial examples are generated from natural model using PGD-$\ell_\infty$ attack ($\epsilon=0.03$). The accuracy is evaluated on the target model. 
    }
    \centering
     \adjustbox{max width=0.7\linewidth}{
    %\vspace{-2mm}
    \begin{tabular}{c|c|c}
        \toprule
        Method &VGG 16& Wide ResNet \\
      \hline
      Adv Train & 79.13\% & 85.84\% \\
      TRADES & 83.53\%& 83.9\% \\
      SPROUT & 86.28\%& 89.1\%\\
      \bottomrule
    \end{tabular}}
    \label{tab:transfer}
    %\vspace{-2mm}
\end{table}

\begin{table}[t]
    %\hspace{-1mm}
    \caption{Accuracy of ImageNet under PGD-$\ell_\infty$ attack }
    \centering
    \adjustbox{max width=1\linewidth}{
    \begin{tabular}{c|c|c|c|c|c}    
    \toprule
        Method & Clean Acc & $\epsilon=0.005$ & $\epsilon=0.01$ & $\epsilon=0.015$ & $\epsilon=0.02$ \\    \hline
        Natural & 78.31\%  & 37.13\% & 9.14\%& 2.12\% & 0.78\%\\
        SPROUT & 74.23\% & 65.24\% & 52.86\% & 35.04\% & 12.18\%\\
    \bottomrule
    \end{tabular}}
    \label{tab:imagenet}
        \hspace{-4mm}
\end{table}

\begin{figure*}[t]
    \centering
    \hspace{-4mm}
    \begin{tabular}{cccc}
        \subfloat[Natural]{\includegraphics[width=0.225\textwidth]{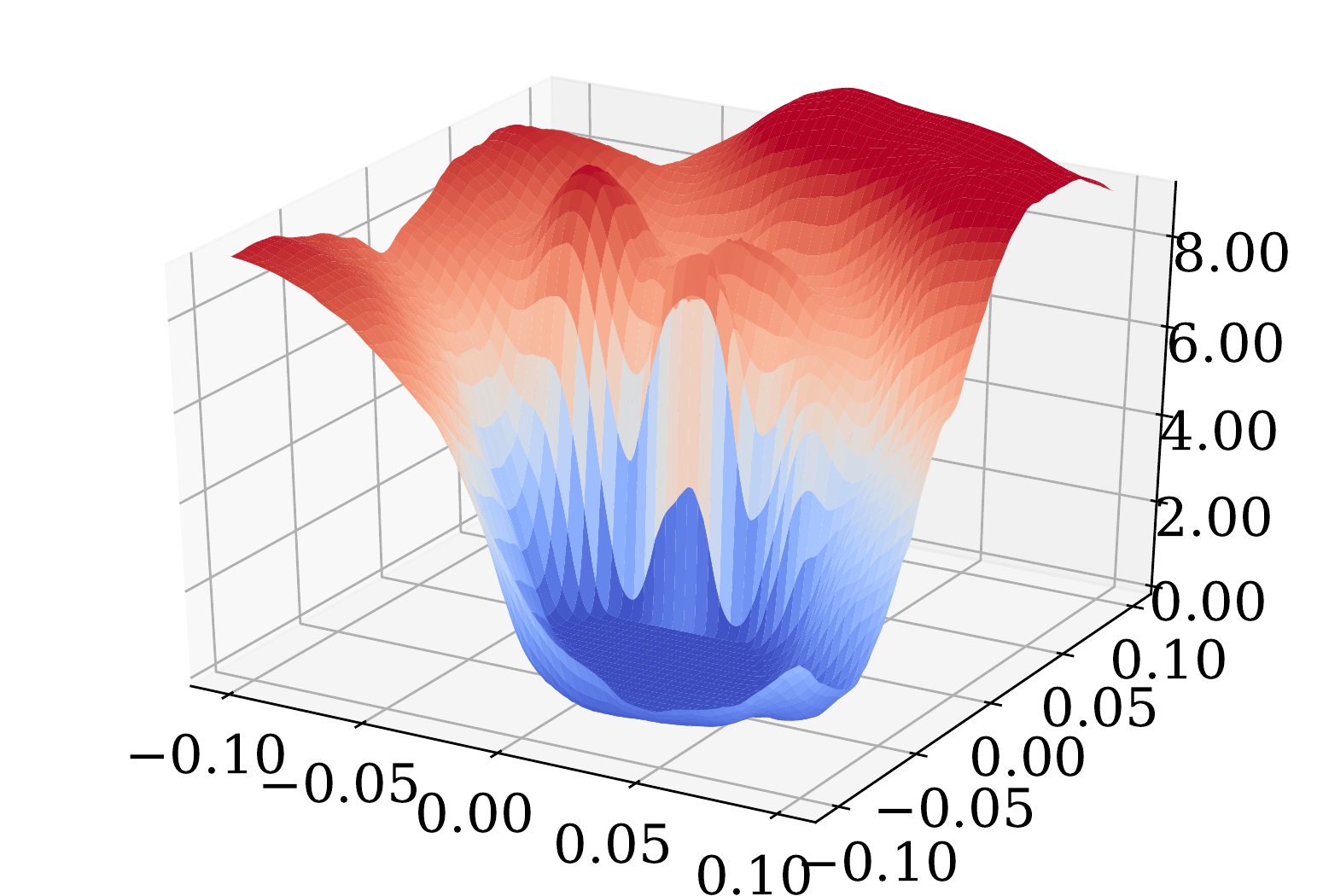}}
		&
		\subfloat[Adv Train]{\includegraphics[width=0.225\textwidth]{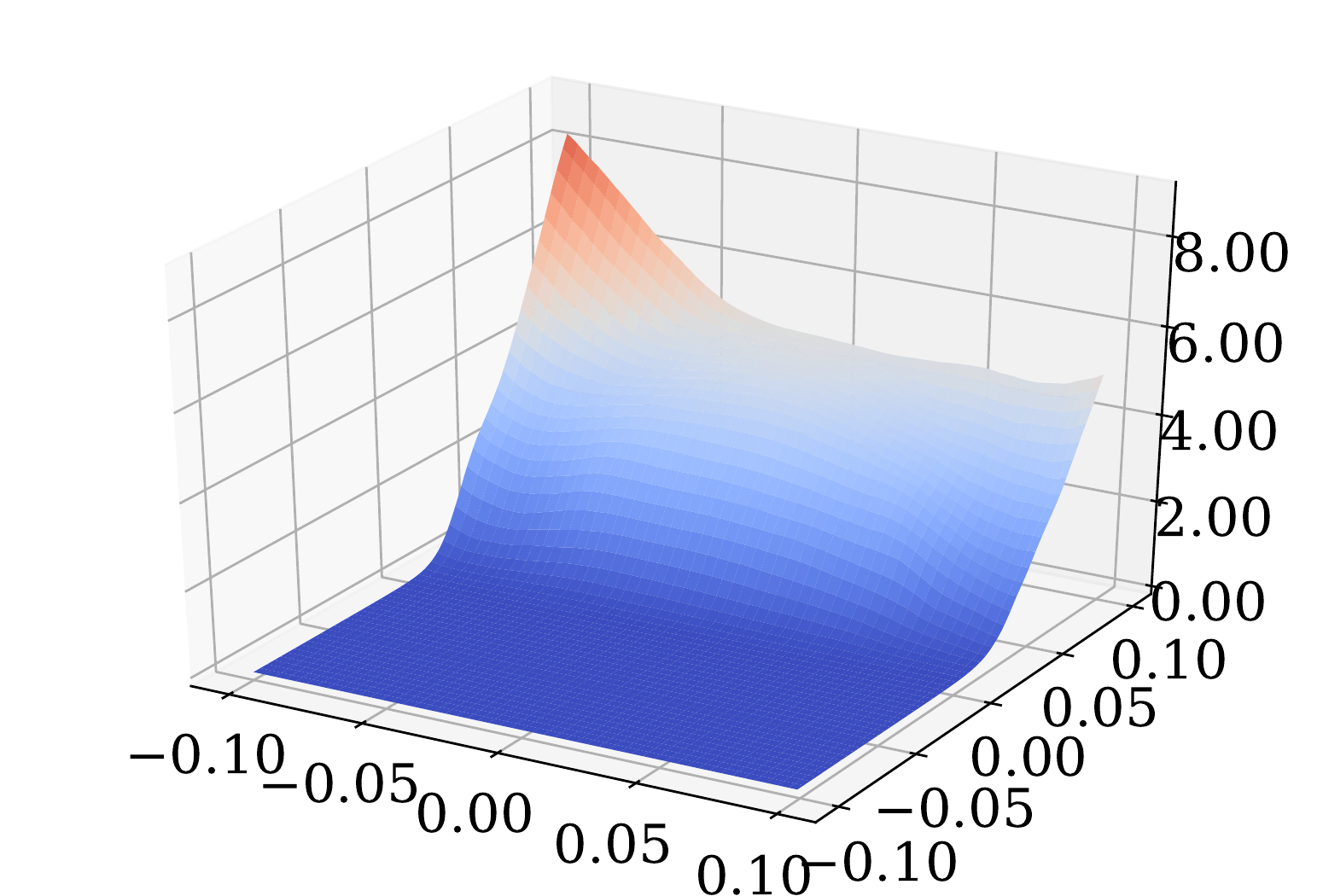}}
		&
        \subfloat[TRADES]{\includegraphics[width=0.225\textwidth]{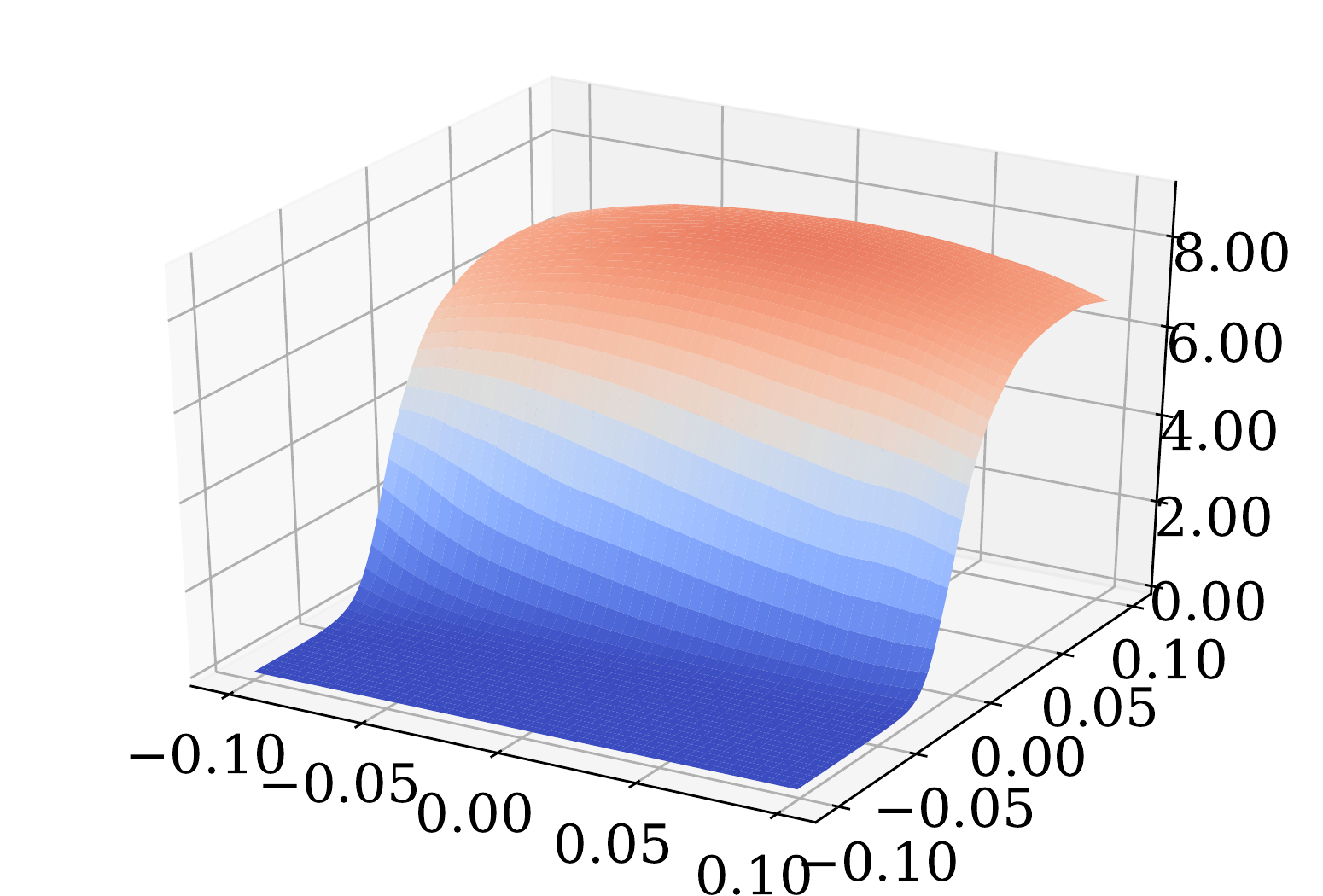}}
		&
		\subfloat[SPROUT]{\includegraphics[width=0.225\textwidth]{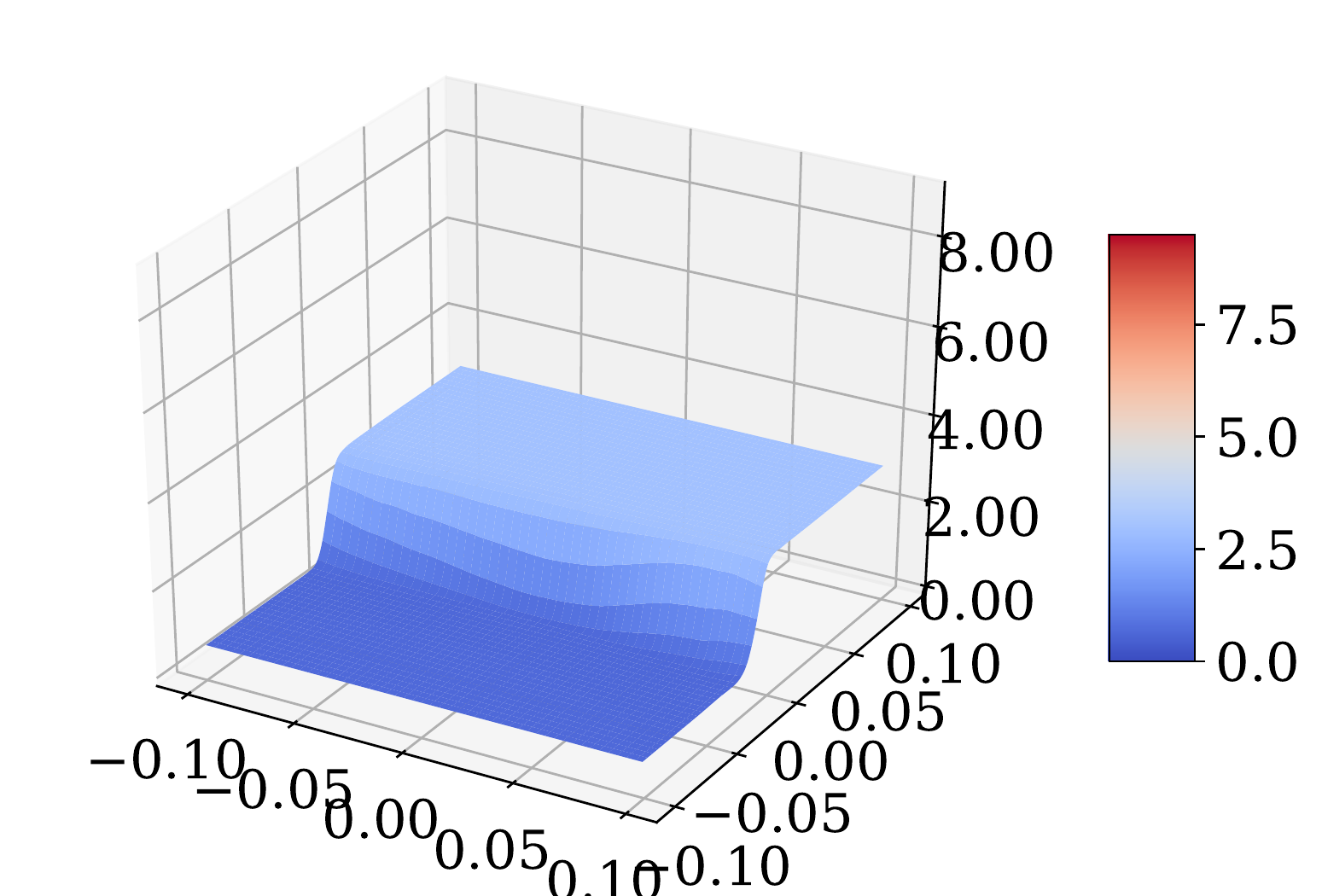}}
	\end{tabular}
% 		\vspace{-2mm}
		\caption{Loss landscape comparison of different training methods}
		\label{fig:loss_landscape}
% 		 \hspace{-4mm}
\end{figure*}

\textbf{ImageNet results.}
As many ImageNet class labels carry similar semantic meanings, to generate meaningful adversarial examples for robustness evaluation, here we follow the same setup as in \citep{athalye2018obfuscated} that adopts PGD-$\ell_\infty$ attacks with randomly targeted labels.
Table~\ref{tab:imagenet} compares the robust accuracy of natural and SPROUT models. SPROUT greatly improves the robust accuracy across different $\epsilon$ values. For example, when $\epsilon=0.01$, SPROUT boosts the robust accuracy of natural model by over $43\%$.
When $\epsilon=0.015 \approx 4/255$, a considerably large adversarial perturbation on ImageNet, SPROUT still attains about $35\%$ robust accuracy while the natural model merely has about $2\%$ robust accuracy. Moreover, comparing the clean accuracy, SPROUT is about 4\% worse than the natural model but is substantially more robust.
We omit the comparison to adversarial training methods as we are unaware of any public pre-trained robust ImageNet models of the same architecture (ResNet-152) prior to the time of our submission, and it is computationally demanding for us to train and fine-tune such large-scale networks with adversarial training.
On our machine, training a natural model takes 31,158.7 seconds and training SPROUT takes 59,201.6 seconds. Comparing to the run-time analysis, SPROUT has a much better scalability than adversarial training and TRADES. However, instead of ResNet-152, we use SPROUT to train the same ResNet-50 model as the pretrained Free Adv Train network and compare their performance in Appendix.

% %\hspace{-2mm}

\subsection{Loss Landscape Exploration}
To further verify the superior robustness using SPROUT,
%Although we have done black-box attack to show our model is not just trying to "block" the gradient to creating a false sense of robustness, 
%to further verify that our robust model is either masking the gradients or making the loss landscape convoluted, 
we visualize the loss landscape of different training methods in Figure \ref{fig:loss_landscape}. Following the implementation in \citep{engstrom2018evaluating}, we vary the data input along a linear space defined by the sign of the input gradient and a random Rademacher vector, where the x- and y- axes represent the magnitude of the perturbation added in each direction and the z-axis represents the loss. One can observe that the loss surface of SPROUT is smoother. Furthermore, it attains smaller loss variation compared with other robust training methods. The results provide strong evidence for the capability of finding more robust models via SPROUT.

% %\vspace{-6mm}

\subsection{Invariance test}
\label{subsection_invariance}
In addition to $\ell_p$-norm bounded adversarial attacks, here we also evaluate model robustness 
against different kinds of input transformations using CIFAR-10 and Wide ResNet. Specifically, we change rotation (with 10 degrees), brightness (increase the brightness factor to 1.5), contrast (increase the contrast factor to 2) and make inputs into grayscale (average all RGB pixel values). 
The model accuracy under these invariance tests is summarized in Table \ref{tab:invariance}. The results show that SPROUT outperforms adversarial training and TRADES. Interestingly,  natural model attains the best accuracy despite the fact that it lacks adversarial robustness, suggesting a potential trade-off between accuracy in these invariance tests and $\ell_p$-norm based adversarial robustness.

\begin{table}[h]
%\vspace{-2mm}
\centering
%\makebox[6pt][c]{\parbox{1\linewidth}{%
\begin{minipage}[b]{1\linewidth}
    \caption{Accuracy under invariance tests}
    \centering
    \adjustbox{max width=1\linewidth}{%
    \begin{tabular}{c|c|c|c|c}
    \toprule
         Method& Rotation & Brightness&  Contrast & Gray\\
    \hline
    Natural & 88.21\% & 93.4\%& 91.88 \% & 91.95\%\\
    Adv Train & 82.66\%&83.64\% &84.99\% & 81.08\%  \\
    TRADES & 80.81\% & 81.5 \%& 83.08\%& 79.27\%\\
    SPROUT & 85.95\% & 88.26 \%& 86.98\%& 81.64\% \\
    \bottomrule
    \end{tabular}}
 %\textcolor{blue}{Missing Adv Training results}}
    \label{tab:invariance}
\end{minipage}
\hfill
\\
%\vspace{-8mm}
% \begin{minipage}[b]{0.52\linewidth}
%     \vspace{-2mm}
%     \caption{Training-time (seconds) for 10 epochs }
%     \centering
%         \vspace{-2mm}
%         \adjustbox{max width=0.95\linewidth}{%
%     \begin{tabular}{c|c|c|c}
%         \toprule
%         \multirow{2}{*}{Methods}& \multicolumn{2}{c}{CIFAR10} & Imagenet\\\cline{2-4}
%         &VGG 16& Wide ResNet & ResNet152 \\
%       \hline
%       Natural  & 146.7 &  1449.6 & 31158.7\\
%       Adv train & 1327.1 & 14246.1 &  $>$1 day\\
%       TRADES & 1932.5& 22438.4 & $>$1 day \\
%       SPROUT & 271.7 & 2727.8 & 59201.6 \\
%       \bottomrule
%     \end{tabular}}
%     \label{tab:scalability}
%     %\vspace{-2mm}
% \end{minipage}
% %}}
% \end{table}
\begin{minipage}[b]{1\linewidth}
    \caption{Training-time (seconds) for 10 epochs }
    \centering
        \adjustbox{max width=0.75\linewidth}{%
    \begin{tabular}{c|c|c}
        \toprule
        \multirow{2}{*}{Methods}& \multicolumn{2}{c}{CIFAR-10} \\\cline{2-3}
        &VGG 16& Wide ResNet\\
      \hline
      Natural  & 146.7 &  1449.6 \\
      Adv Train & 1327.1 & 14246.1 \\
      TRADES & 1932.5& 22438.4 \\
      SPROUT & 271.7 & 2727.8\\
      Free Adv Train(m=8) & 2053.1 & 20652.5\\
      \bottomrule
    \end{tabular}}
    \label{tab:scalability}
    %\vspace{-2mm}
\end{minipage}
%}}
\end{table}

\subsection{Scalability}
\label{subsec_scalability}
SPROUT enjoys great scalability over adversarial training based algorithms because its training requires much less number of back-propagations per iteration, which is a dominating factor that contributes to considerable run-time in adversarial training.
Table \ref{tab:scalability} benchmarks the run-time of different training methods for 10 epochs. On CIFAR-10, the run-time of adverarial training and TRADES is about 5$\times$ more than SPROUT. We also report the run-time analysis using the default implementation of Free Adv Train \citep{shafahi2019adversarial}. Its 10-epoch run-time with the replay parameter $m=8$ is similar to TRADES. But we also note that Free Adv Train may require less number of epochs when training to convergence.

% % \begin{table}[h]
% % \vspace{-2mm}
% %     \caption{Run-time (seconds) for training 10 epochs }
% %     \centering
% %     \vspace{-2mm}
% %     \begin{tabular}{c|c|c|c}
% %         \toprule
% %         \multirow{2}{*}{Methods}& \multicolumn{2}{c}{CIFAR10} & Imagenet\\\cline{2-4}
% %         &VGG 16& Wide ResNet & ResNet152 \\
% %       \hline
% %       Natural  & 146.7 &  1449.6 & \\
% %       Adv train & 1327.1 & 14246.1 &  \\
% %       TRADES & 1932.5& 22438.4 & \\
% %       SPROUT & 271.7 & 2727.8 & 59201.6 \\
% %       \bottomrule
% %     \end{tabular}
% %     \label{tab:scalability}
% %     \vspace{-2mm}
% % \end{table}

\subsection{Ablation Study}
\label{subsec_ablation}
\textbf{Dissecting SPROUT.} Here we perform an ablation study using VGG-16 and CIFAR-10 to 
investigate and factorize the robustness gain in SPROUT's  three modules: Dirichlet label smoothing (Dirichlet), Gaussian augmentation (GA) and Mixup. We implement all combinations of these techniques and include uniform label smoothing (LS) \citep{szegedy2016rethinking} as another baseline. Their accuracies under PGD-$\ell_\infty$ 0.03 attack are shown in Table \ref{tab:aba}. We highlight some important findings as follows. 
\\
$\bullet$ Dirichlet outperforms uniform LS by a significant factor, suggesting the importance of our proposed self-progressing label smoothing in improving adversarial robustness.
 \\ 
 $\bullet$ Comparing the performance of individual modules alone (GA, Mixup and Dirichlet), our proposed Dirichlet attains the best robust accuracy, suggesting its crucial role in training robust models.
 \\
 $\bullet$ No other combinations can outperform SPROUT. Moreover, the robust gains from GA, Mixup and Dirichlet appear to be \textit{complementary}, as SPROUT's accuracy is close to the sum of their individual accuracy. To justify their diversity in robustness, we compute the cosine similarity of their pairwise input gradients and find that they are indeed quite diverse and thus can promote robustness when used together. The details are given in Appendix.

\begin{table}
    \caption{Robust accuracy under $\ell_\infty$ 0.03 strength with different combinations of the modules in SPROUT.}
    \centering
        \adjustbox{max width=0.75\linewidth}{%
    \begin{tabular}{c|c|c}
        \toprule
        \multirow{2}{*}{Methods}& \multicolumn{2}{c}{VGG 16} \\\cline{2-3}
        &PGD$^{20}$& PGD$^{100}$\\
      \hline
      GA  & 12.44\% & 9.46\%   \\
      Mixup & 0.76\% & 0.08\% \\
      Dirichlet & 29.64\%& 9.77\% \\
      GA+Mixup & 41.88\% & 40.29\%\\
      Mixup+Dirichlet & 17.53\%& 7.97\%\\
      GA+Dirichlet & 55.27\%& 53.79\%\\
      Uniform LS & 15.36\%& 5.12\% \\
      SPROUT &62.64\% & 58.93\%\\
      \bottomrule
    \end{tabular}}
    \label{tab:aba}
    %\vspace{-2mm}
%}}
\end{table}

% \begin{figure}[t]
% 	%\vspace{-2mm}
%     \centering
%         \adjustbox{max width=1\linewidth}{
%     \begin{tabular}{cc}
%     \subfloat{\includegraphics[width=0.38\textwidth]{imgs/vgg_20_aba.pdf}}
% 		&
% 		\subfloat{\includegraphics[width=0.54\textwidth]{imgs/vgg_100_aba.pdf}}
%         \\ 
% 	\end{tabular}}
% 	\vspace{-2mm}
% 		\caption{Robust accuracy with different combinations of the modules in SPROUT  }%\textcolor{blue}{change the legend of LS to uniform LS. Also need to move legend outside the plot}}
% 		\label{fig:aba}
% % 			\vspace{-2mm}
% \end{figure}

\begin{figure}[t]
    \centering
    \adjustbox{max width=1\linewidth}{
\includegraphics[width=0.35\textwidth]{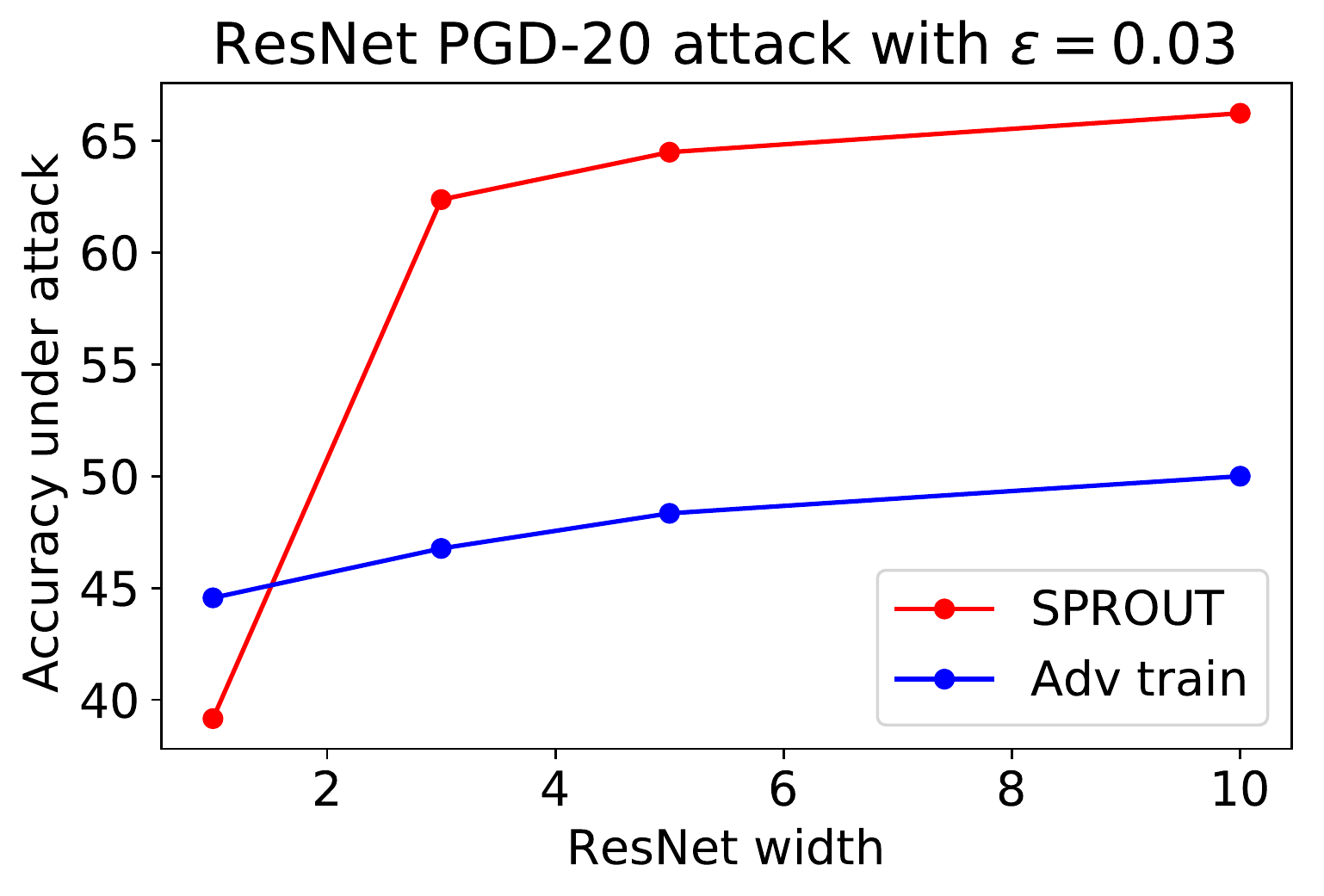}}

% 		\vspace{-4mm}
		\caption {Effect of network width against PGD-$\ell_\infty$ attack
		on CIFAR-10 and ResNet-34.}
		\label{fig:width}
% 		\vspace{-2mm}
\end{figure}

\textbf{Effect on network width.}
It was shown in \citep{madry2017towards} that adversarial training (Adv Train) will take effect when a network has sufficient capacity, which can be achieved by increasing network width. Figure \ref{fig:width} compares the robust accuracy of SPROUT and Adv Train with varying network width on Wide ResNet and CIFAR-10. When the network has width = 1 (i.e. a standard ResNet-34 network \citep{he2016deep}), the robust accuracy of SPROUT and Adv Train are both relatively low (less than 47\%). However, as the width increases, SPROUT soon attains significantly better robust accuracy than Adv Train by a large margin (roughly 15\%).
Since SPROUT is more effective in boosting robust accuracy as network width varies, the results also suggest that SPROUT can better support robust training for a broader range of network structures.

%\PY{a figure of SPROUT with different width}
%\PY{a table compating to adv training with wideth = 1,3,5,10}

%\textcolor{blue}{performance of SPROUT vs varying $\alpha$ or $\lambda$} (if time permits)
	%\vspace{-3mm}
% \begin{figure}[t]
%     \centering
%     \includegraphics[width=0.32\textwidth]{imgs/width.pdf}
%     %\begin{tabular}{cc}
% %     \subfloat[]{\includegraphics[width=0.45\textwidth]{imgs/width.pdf}}
% % 		&
% % 		\subfloat[Sensitivity of   $\lambda$ and $\alpha$ in SPROUT \label{fig:para}]{\includegraphics[width=0.45\textwidth]{imgs/vgg_20_para.pdf}}
% %         \\ 
% %	\end{tabular}
% % 		\vspace{-2mm}
% 		\caption{Effect of network width against PGD-$\ell_\infty$ attack
% 		on CIFAR-10 and ResNet-34.  }
% 		\label{fig:width}
% % 			\vspace{-2mm}
% \end{figure}

\section{Conclusion}
This paper introduced SPROUT, a self-progressing robust training method motivated by vicinity risk minimization. 
When compared with state-of-the-art adversarial training based methods, our extensive experiments showed that the proposed self-progressing Dirichlet label smoothing technique in SPROUT plays a crucial role in finding substantially more robust models  against $\ell_\infty$-norm bounded PGD attacks and simultaneously makes the corresponding model more generalizable to various invariance tests. We also find that SPROUT can strengthen a wider range of network structures as it is less sensitive to network width changes. Moreover, SPOURT's self-adjusted learning methodology not only makes its training free of attack generation but also becomes scalable solutions to large networks. Our results shed new insights on devising  comprehensive and robust training methods that are attack-independent and scalable.

\section{Acknowledgments}
This work was done during Minhao Cheng's internship at IBM Research. Cho-Jui Hsieh and Minhao Cheng are partially supported by National Science Foundation (NSF) under IIS-1901527, IIS-2008173 and Army Research Lab under W911NF-20-2-0158. 

\bibliography{icml2020}

\clearpage
\appendix
\section{Appendix}

\subsection{Exact Performance Metrics for Figure \ref{fig:radar}}
\label{appen_radar}

The performance metrics of Figure \ref{fig:radar} are shown in Table \ref{tab:overall}.

\begin{table*}[t]
    \caption{Performance comparison between different training methods on  VGG-16 and CIFAR-10}
    \centering
      \adjustbox{max width=1\linewidth}{
    \begin{tabular}{c|c|c|c|c|c}
    \toprule
        Method & Clean Acc & $\ell_\infty$ Acc ($\epsilon=0.03$) & C\&W Acc & Invariance (Contrast) & Scalibility (10 epochs) \\
        \hline        
        Natural & 95.93\% &	0\%	& 26.95\% &	91.88\% &	146.7 (secs)	\\
        Adv Train&  84.92\% & 36.29\%& 70.13\% & 84.99\%&1327.1 (secs)\\
        TRADES & 88.6\% & 38.29\%&75.08\%&	83.08\% &1932.5 (secs)\\
        SPROUT & 90.56\% &58.93\%&72.7\%& 86.98\%&271.7 (secs)\\
        \bottomrule
    \end{tabular}
    }
    \label{tab:overall}
\end{table*}

\subsection{Learned Label Correlation from SPROUT}
\label{appen_label_corr}
\begin{figure*}[h]
    \centering
    \includegraphics[width=0.7\textwidth]{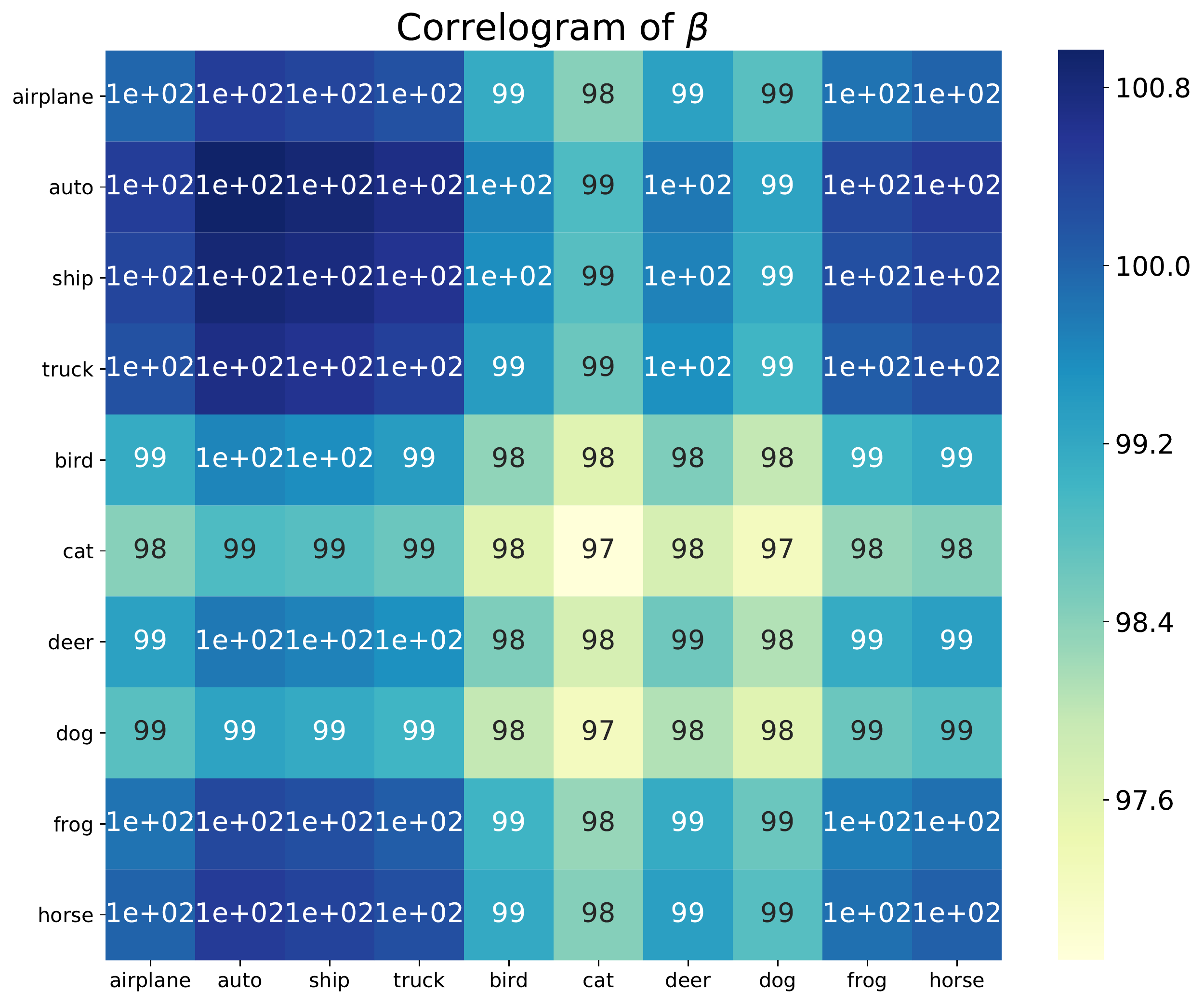}
    \caption{Matrix plot of the product $\beta_s \cdot \beta_t$ of the learned $\bbeta$ parameter on CIFAR-10 and VGG-16.}
    \label{fig:label_corr}
\end{figure*}

Based on the statistical properties of Dirichlet distribution in \eqref{eqn_corr}, we use the final $\bbeta$ parameter learned from Algorithm \ref{alg:sprout} with CIFAR-10 and VGG-16 to display the matrix of its pair-wise product $\beta_s \cdot \beta_t$ in Figure \ref{fig:label_corr}. The value in each entry is proportional to the absolute value of the label covariance in \eqref{eqn_corr}. We observe some clustering effect of class labels in CIFAR-10, such as relatively high values among the group of \{airplane, auto, ship, truck\} and relatively low values among the group of \{bird, cat, deer, dog\}. Moreover, since the $\bbeta$ parameter is progressively adjusted and co-trained during model training, and the final $\bbeta$ parameter is clearly not uniformly distributed, the results also validate the importance of using parametrized label smoothing to learn to improve robustness.

\subsection{Parameter Sensitivity Analysis}
\label{appen_para}
\begin{figure}[htpb]
    \centering	\includegraphics[width=0.45\textwidth]{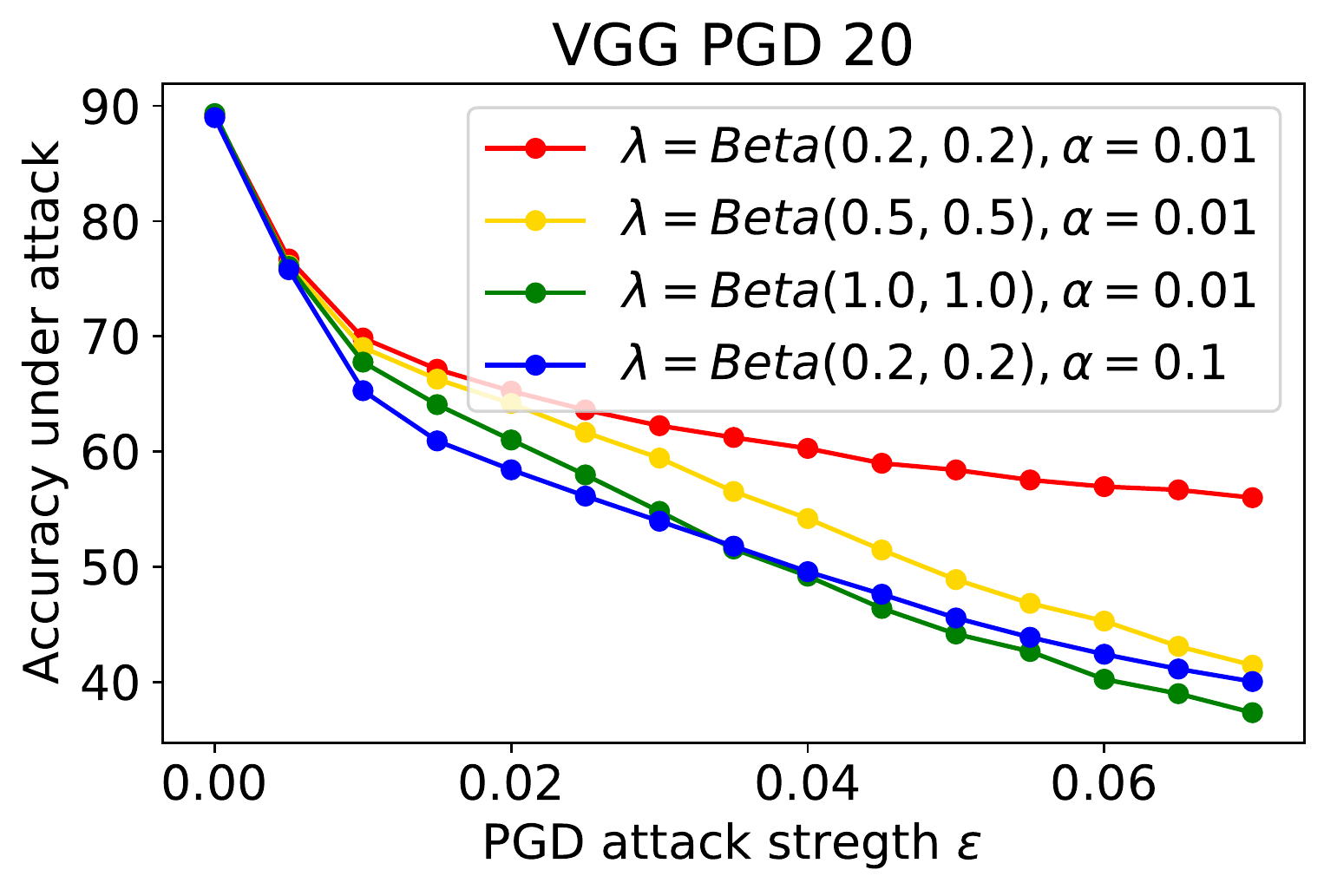}
		%\vspace{-2mm}
		\caption{Sensitivity of hyperparameters  $\lambda$ and $\alpha$ in SPROUT under PGD-$\ell_\infty$ attack}
	\label{fig:para}
\end{figure}

We perform an sensitivity analysis of the mixing parameter $\lambda \sim$ Beta($a$,$a$) and the smoothing parameter $\alpha$ of SPROUT in Figure \ref{fig:para}. When fixing $a$, we find that setting $\alpha$ too large may affect robust accuracy, as the resulting training label distribution could be too uncertain to train a robust model. Similarly, when fixing $\alpha$, setting $a$ too large may also affect robust accuracy.

%\textcolor{blue}{performance of SPROUT vs varying $\alpha$ or $\lambda$} (if time permits)

\subsection{Performance with different number of random starts for PGD attack}
\label{appen_PGD_multi}
As suggested by \citep{madry2017towards}, PGD attack with multiple random starts is a stronger attack method to evaluate robustness. Therefore, in Table~\ref{tab:start}, we conduct the following experiment on CIFAR-10 and Wide ResNet and find that the model trained by SPROUT can still attain at least 61\% accuracy against PGD-$\ell_{\infty}$ attack ($\epsilon=0.03$) with the number of random starts varying from 1 to 10 and with 20 attack iterations. On the other hand, the accuracy of Adversarial training and TRADES can be as low as 45.21\% and 56.7\%, respectively. Therefore,
The robust accuracy of SPROUT is still clearly higher than other methods. We can conclude that increasing the number of random starts may further reduce the robust accuracy of all methods by a small margin, but the observed robustness rankings and trends among all methods are unchanged. We also perform one additional attack setting: 100-step PGD-$\ell_{\infty}$ attack with 10 random restarts and $\epsilon=0.03$. We find that SPROUT can still achieve 61.18\% robust accuracy.

% We also perform two additional attack settings: (i)
% 100-step PGD-$\ell_{\infty}$ attack with 10 random restarts using the C\&W loss and $\epsilon=0.03$; (ii) 100-step PGD-$\ell_{\infty}$ attack with 10 random restarts using the cross entropy and $\epsilon=0.03$. We find that SPROUT can still achieve 51.23\% robust accuracy in setting (i) and  61.18\% robust accuracy in setting (ii).

\begin{table*}[htbp]
    \caption{Robust accuracy of different training methods under  PGD-$\ell_\infty$ attack with $\epsilon=0.03$ using different number of random starts}
    \label{tab:start}
    \centering
    \begin{tabular}{c|c|c|c|c|c}
    \toprule
        $\#$ random start &1 & 3 & 5 & 8 & 10 \\
         \hline
        Adversarial training & 45.88\% & 45.67\%& 45.52\% & 45.52\%  & 45.21\% \\
        TRADES & 57.02\% & 56.84\% & 56.77\% & 56.7\% & 56.7\%\\
        SRPOUT & 64.58\% & 62.53\% & 61.98\% & 61.38\%& 61.00\%\\
    \bottomrule
    \end{tabular}
\end{table*}

\subsection{Performance on C\&W-$\ell_\infty$ attack}
\label{appen_CW_Linf}
\begin{figure}
    \centering
    \includegraphics[width=0.45\textwidth]{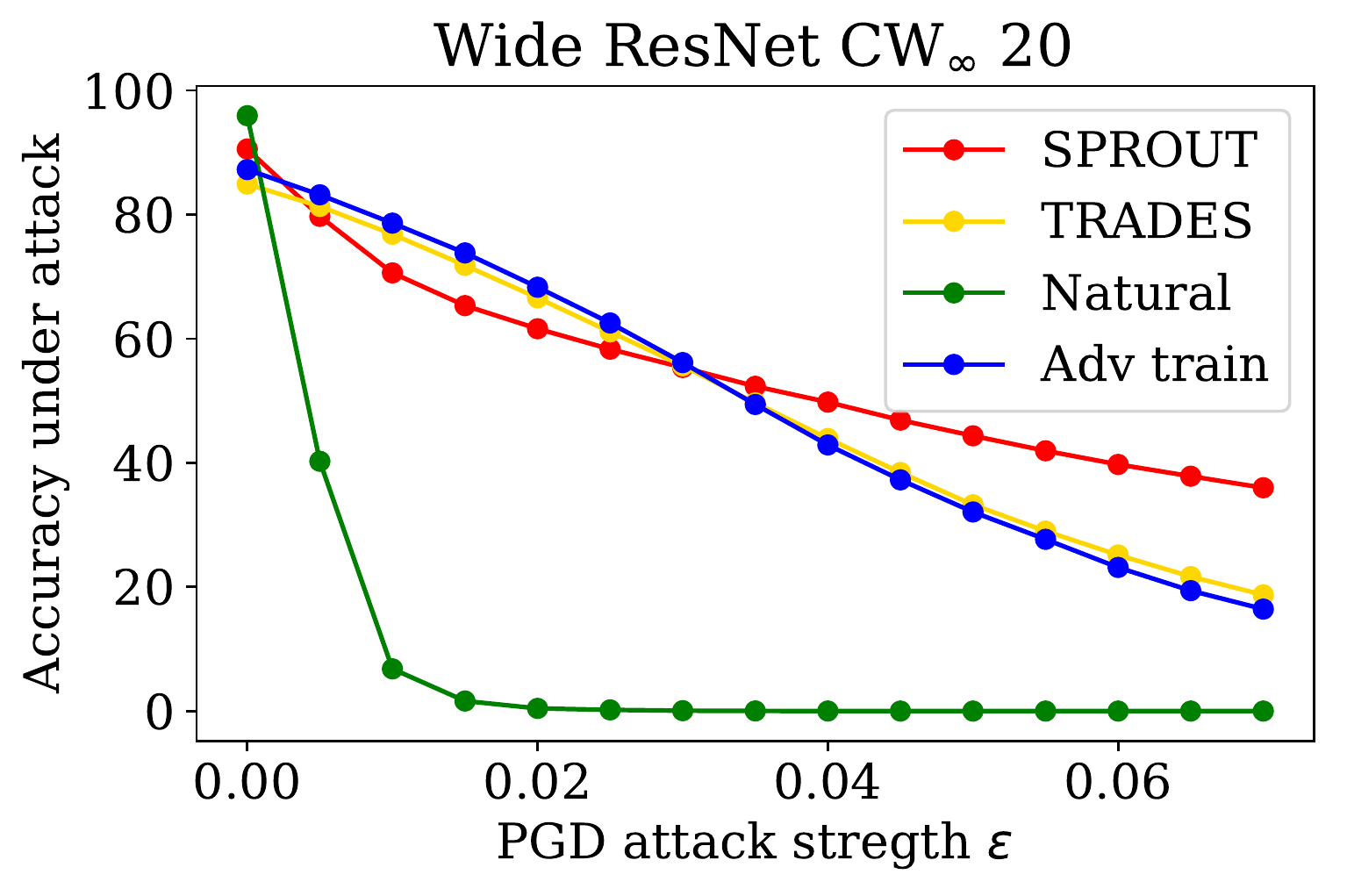}
    \caption{Robust accuracy under C\&W-$\ell_\infty$ attack. 
    %SPROUT outperforms other methods when $\epsilon\geq0.03$.
    }
    \label{fig:cw_inf}
\end{figure}

To further test the robustness on $\ell_\infty$ constraint, we replace the cross entropy loss with C\&W-$\ell_\infty$ loss \citep{carlini2017towards} in PGD attack. Similar to the PGD-$\ell_\infty$ attack results,
Figure~\ref{fig:cw_inf} shows that although SPROUT has slightly worse accuracy under small $\epsilon$ values, it attains much higher robust accuracy when $\epsilon\geq0.03$.

\subsection{Performance comparison with Free Adversarial training on ResNet-50 and ImageNet}
\label{appen_resnet}

Here we compare the performance of SPROUT with a pre-trained robust ResNet-50 model on ImageNet, which is shared by the authors in \citep{shafahi2019adversarial} proposing the free adversarial training method (Free Adv Train). We find that SPROUT obtains similar robust accuracy as Free Adv Train when $\epsilon \leq 0.01$. As $\epsilon$ becomes larger, Free Adv Train has better accuracy. However, comparing to the performance of ResNet-152 in Table \ref{tab:imagenet}, SPROUT's clean accuracy on ResNet-50 actually drops by roughly 13\%, indicating a large performance gap that intuitively shound not be present.
Therefore, based on the current results, we postulate that the training parameters of SRPOUT for ResNet-50 may not have been fully optimized (we use the default training parameters of ResNet-152 for ResNet-50), and that it is possible that SPROUT has larger gains in robust accuracy as the RestNet models become deeper.
%However, we also note that the clean accuracy of SPROUT is more than 13.7\% higher than that of Free Adv train. The robust gain of Free Adv train for large  $\epsilon$ values can be a consequence of sacrificing notable clean accuracy to trade for robustness. Therefore, without assuring similar clean accuracy, the robustness of these two models may not be directly comparable. 

\begin{table*}[htbp]
 %   \hspace{-1mm}
    \caption{Robust accuracy under PGD-$\ell_\infty$ random targeted attack on ImageNet and ResNet-50}
    \centering
    \adjustbox{max width=1\linewidth}{
    \begin{tabular}{c|c|c|c|c|c}    
    \toprule
        Method & Clean Acc & $\epsilon=0.005$ & $\epsilon=0.01$ & $\epsilon=0.015$ & $\epsilon=0.02$ \\    \hline
        Natural & 76.15\%& 24.37\%& 3.54\%& 0.90\%&0.40\%\\
        Free Adv Train& 60.49\%&51.35\%& 42.29\%& 32.96\%&24.45\% \\
        SPROUT& 61.23\%&51.69\%&38.14\%&25.98\%&18.52\%\\
    \bottomrule
    \end{tabular}}
    \label{tab:imagenet-50}
        %\hspace{-4mm}
\end{table*}

\begin{table}[htbp]
    \centering
    \caption{Average pair-wise cosine similarity of the three modules in SPROUT}
    \begin{tabular}{c|c|c|c}
    \toprule
         &Dirichilet LS& Mixup & GA \\
         \hline
        Dirichilet LS & NA & 0.1023 & 0.0163\\
        Mixup &0.1023 &NA & 0.0111 \\
        GA &0.0163 &0.0111 &NA \\
    \bottomrule
    \end{tabular}
    \label{tab:cos}
\end{table}

\subsection{Diversity Analysis}
\label{appen_diversity}
%\textcolor{blue}{should this be cosine similarity of cosine distance?}
In order to show the three modules (Dirichlet LS, GA and Mixup) in SPROUT lead to robustness gains that are complementary to each other,
%does learn the model with different adversarial robustness property, 
we perform a diversity analysis motivated by \citep{kariyappa2019improving} to
measure the similarity of their pair-wise input gradients and report the average cosine similarity in Table~\ref{tab:cos} over 10,000 data samples using CIFAR-10 and VGG-16. 
%\textcolor{blue}{what is cosine distance, need a definition}
We find that the pair-wise similarity between modules is  indeed quite small ($<0.103$).
%and is comparable to the cosine similarity of independently drawn standard normal random vectors.
The Mixup-GA similarity is the smallest among all pairs since the former performs both label and data augmentation based on convex combinations of training data, whereas the latter only considers random data augmentation.
The Dirichlet\_LS-GA similarity is the second smallest (and it is close to the Mixup-GA similarity) since the former progressively adjusts the training label $\tilde{\by}$ while the latter only randomly adjusts the training sample $\tilde{\bx}$.  The Dirichlet\_LS-Mixup similarity is relatively high because Mixup depends on the training samples and their labels while Dirichlet LS also depend on them and the model weights.
The results show that their input gradients are diverse as they point to vastly different directions. Therefore, SPROUT enjoys complementary robustness gain and can promote robustness when combining these techniques together.

\subsection{PGD attacks with more iterations}
\begin{figure}[htbp]
    \centering
    \includegraphics[width=0.4\textwidth]{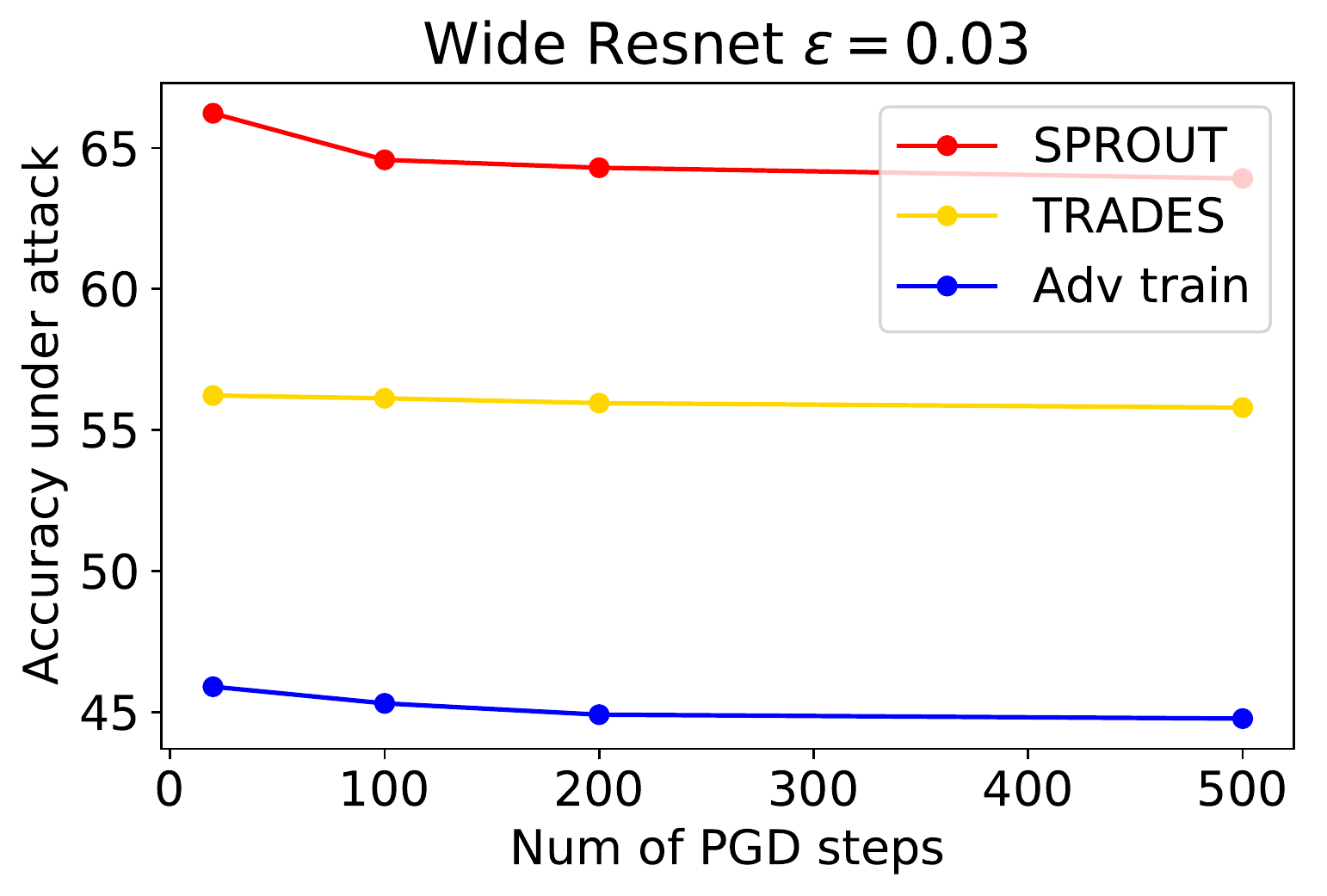}
    \caption{Stability in PGD-$\ell_\infty$ attack with different PGD steps}
    \label{fig:num_aba}
\end{figure}

To ensure the robustness of SPROUT is not an artifact of running insufficient iterations in PGD attack \citep{athalye2018obfuscated}, 
Figure \ref{fig:num_aba} shows the robust accuracy with varing number of PGD-$\ell_\infty$ attack steps from 10 to 500 on Wide ResNet and CIFAR-10. The results show stable performance in all training methods once the number of attack steps exceeds 100. It is clear that SPROUT indeed outperforms others by a large margin.

\subsection{Model weight initialization.}
\begin{figure}[htbp]
    \centering
    \includegraphics[width=0.5\textwidth]{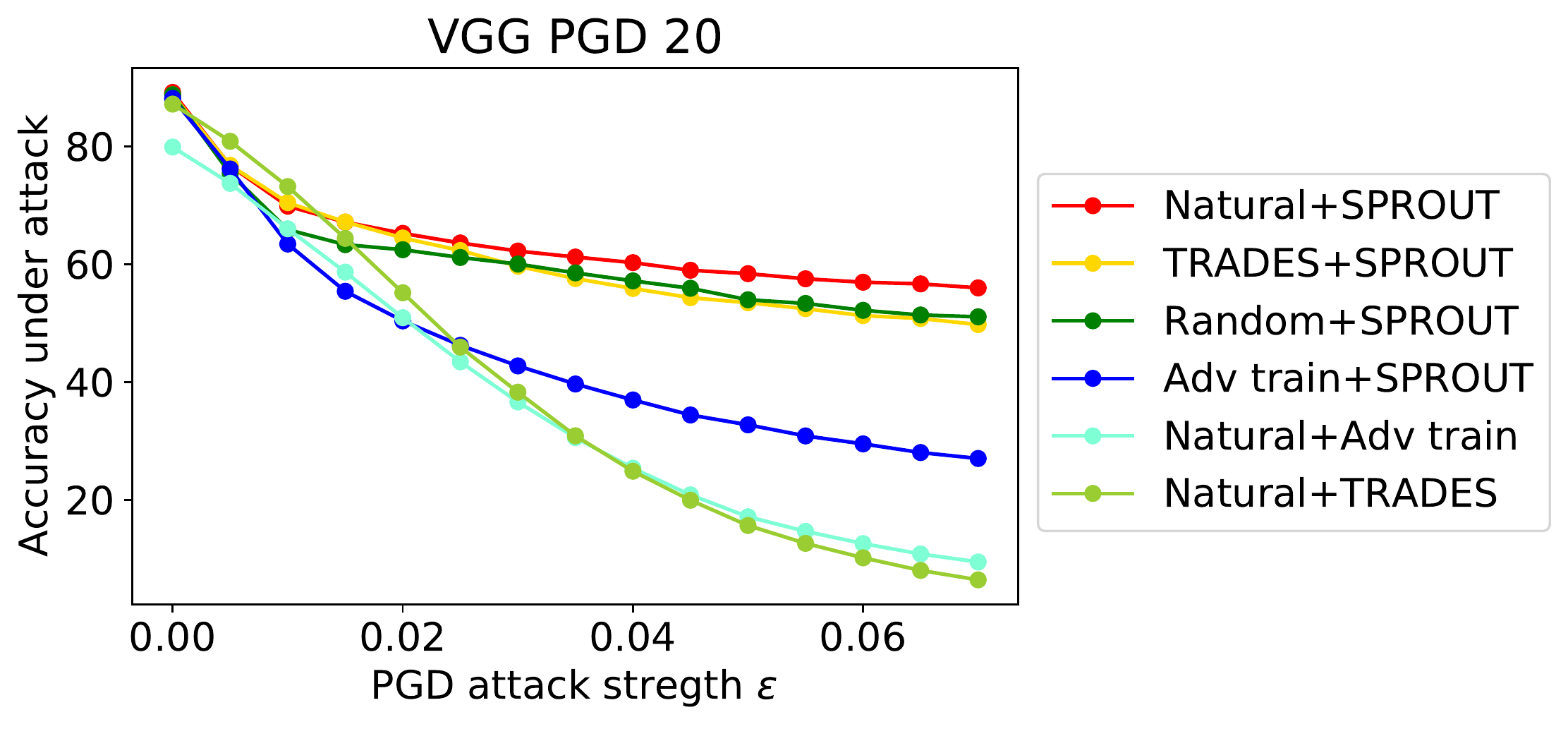}
    \caption{Stability in PGD-$\ell_\infty$ attack with different PGD steps}
    \label{fig:resume_aba}
\end{figure}

Figure \ref{fig:resume_aba} compares the effect of model initialization using CIFAR-10 and VGG-16 under PGD-$\ell_\infty$ attack, where the legend $A+B$ means using Model $A$ as the initialization and training with Method $B$. Interestingly, Natural+SPROUT attains the best robust accuracy when $\epsilon \geq 0.02$. TRADES+SPROUT and Random+SPROUT also exhibit strong robustness since their training objective involves the loss on both clean and adversarial examples. In contrast, Adv Train+SPROUT does not have such benefit since adversarial training only aims to minimize adversarial loss. This finding is also unique to SPROUT, as neither Natural+Adv Train nor Natural+TRADES can boost robust accuracy. Our results provide novel perspectives on improving robustness and also
indicate that SPROUT is indeed a new robust training method that vastly differs from adversarial training based methods.

Moreover, SPROUT performs better when initializing with natural pre-trained model. Therefore, in Figure \ref{fig:resume_aba}, we have tried different kinds of initialization such as random, adversarial training and TRADES.

% \begin{figure}[h]
%     \centering
%     \adjustbox{max width=1\linewidth}{
%     \begin{tabular}{c}
%     \subfloat[\label{fig:num_aba}]{\includegraphics[width=0.6\textwidth]{imgs/num_aba.pdf}}
% 		&\\
% 		\subfloat[ \label{fig:resume_aba}]{\includegraphics[width=0.8\textwidth]{imgs/resume_aba.pdf}}
%         \\ 
% 	\end{tabular}}
% 		\vspace{-4mm}
% 		\caption{Stability in PGD-$\ell_\infty$ attack and the effect of model initialization. Left: (a) Robust accuracy with different PGD steps. Right: (b) Robust accuracy with different model initialization.}
% \end{figure}

\end{document}